\documentclass[11pt,oneside,english,british]{amsbook}
\usepackage{charter}

\usepackage[T1]{fontenc}
\usepackage[latin9]{inputenc}
\usepackage{geometry}
\usepackage{color}
\definecolor{document_fontcolor}{rgb}{0.0390625, 0, 0.109375}
\color{document_fontcolor}
\usepackage{babel}
\usepackage{array}
\usepackage{float}
\usepackage{multirow}
\usepackage{amsthm}
\usepackage{amsbsy}
\usepackage{amstext}
\usepackage{amssymb}
\usepackage{graphicx}
\usepackage{setspace}
\usepackage{esint}
\usepackage[unicode=true,
 bookmarks=true,bookmarksnumbered=false,bookmarksopen=false,
 breaklinks=false,pdfborder={0 0 1},backref=false,colorlinks=false]
 {hyperref}
\hypersetup{pdftitle={Feature Normalisation for Robust Speech Recognition},
 pdfauthor={D S Pavan Kumar},
 pdfsubject={MS Thesis, Indian Institute of Technology Madras}}

\makeatletter

\providecommand{\tabularnewline}{\\}

\numberwithin{section}{chapter}
\numberwithin{equation}{section}
\numberwithin{figure}{section}
\newenvironment{lyxlist}[1]
{\begin{list}{}
{\settowidth{\labelwidth}{#1}
 \setlength{\leftmargin}{\labelwidth}
 \addtolength{\leftmargin}{\labelsep}
 }}
{\end{list}}

\usepackage{subfig,bbm,breakcites}
\@ifundefined{showcaptionsetup}{}{%
 \PassOptionsToPackage{caption=false}{subfig}}
\usepackage{subfig}
\makeatother

\begin{document}

\title{Feature Normalisation for Robust Speech Recognition}

\author{D. S. Pavan Kumar}

\date{October 2013}

\maketitle
\begin{center}
\emph{\footnotesize{}This thesis was submitted in October 2013 for
the}\\
\emph{\footnotesize{}award of the degree Master of Science by Research,}\\
\emph{\footnotesize{} under the supervision of Prof. S. Umesh,}\\
\emph{\footnotesize{}at the Department of Electrical Engineering,}\\
\emph{\footnotesize{}Indian Institute of Technology Madras, Chennai.}
\par\end{center}{\footnotesize \par}

\begin{center}
\emph{\bigskip{}
}
\par\end{center}
\begin{abstract}
Speech recognition system performance degrades in noisy environments.
If the acoustic models (HMMs) for speech are built using features
of clean utterances, the features of a noisy test utterance would
be acoustically mismatched with the trained model. This gives poor
likelihood values and poor recognition accuracy. Model adaptation
and feature normalisation are two broad areas that address this problem.
While the former often gives better performance, the latter involves
estimation of lesser number of parameters, making the system feasible
for practical implementations.

This research focuses on the efficacies of various subspace, statistical
and stereo based feature normalisation techniques. A subspace projection
based method has been investigated as a standalone and adjunct technique
involving reconstruction of noisy speech features from a precomputed
set of clean speech building-blocks. The building blocks are learned
using non-negative matrix factorisation (NMF) on log-Mel filter bank
coefficients, which form a basis for the clean speech subspace. The
work provides a detailed study on how the method can be incorporated
into the extraction process of Mel-frequency cepstral coefficients.
Experimental results show that the new features are more robust to
noise, and achieve better results when combined with the existing
techniques.

The work also proposes a modification to the training process of the
popular SPLICE algorithm for noise robust speech recognition. The
modification is based on feature correlations, and enables this stereo-based
algorithm to improve the performance in all noise conditions, especially
in unseen cases. Further, the modified framework is extended to work
for non-stereo datasets where clean and noisy training utterances,
but not stereo counterparts, are required. An MLLR-based computationally
efficient run-time noise adaptation method in SPLICE framework has
been proposed.
\end{abstract}

\tableofcontents{}

\chapter*{Abbreviations and Notations}
\begin{lyxlist}{00.00.0000}
\begin{singlespace}
\item [{ASR}] Automatic Speech Recognition\end{singlespace}

\item [{CMS}] Cepstral Mean Subtraction
\item [{CMVN}] Cepstral Mean and Variance Normalisation
\item [{GMM}] Gaussian Mixture Model
\item [{HEQ}] Histogram Equalisation
\item [{HLDA}] Heteroscedastic Linear Discriminant Analysis
\item [{HMM}] Hidden Markov Model
\item [{IVR}] Interactive voice response
\item [{LMFB}] Log-Mel Filterbank
\item [{MFCC}] Mel-Frequency Cepstral Coefficient
\item [{MLLR}] Maximum Likelihood Linear Regression
\item [{NMF}] Non-Negative Matrix Factorisation
\item [{PCA}] Principal Component Analysis
\item [{SPLICE}] Stereo-Based Piecewise Linear Compensation for Environments
\item [{VQ}] Vector Quantisation
\item [{\bigskip{}
}]~
\item [{$\boldsymbol{\mu}$}] Mean vector
\item [{$\boldsymbol{\Sigma}$}] Covariance matrix
\item [{$\mathcal{D}$}] Divergence function
\item [{$\mathbf{\boldsymbol{\mathcal{H}}}$}] HLDA matrix
\item [{$\mathbf{\mathcal{L}}$}] Likelihood function
\item [{$\mathbb{R}^{D+}$}] Positive quadrant of the $D-$dimensional
real number space
\item [{$T$}] (Normal caps font) Constant
\item [{$t$}] (Normal small font) Variable to the corresponding normal
caps font notation
\item [{$\mathbf{V}$}] (Bold caps font) Matrix
\item [{$\mathbf{x}$}] (Bold small font) Vector
\item [{$\left\{ \mathbf{x}_{n}\right\} $}] Set of vectors $\mathbf{x}_{1},\mathbf{x}_{2},\ldots,\mathbf{x}_{N}$
\item [{$\otimes$}] Hadamard (element-wise) matrix product
\item [{$\left[\text{·}\right]_{dn}$}] Element of the $d^{th}$ row, $n^{th}$
column of the matrix
\end{lyxlist}

\chapter{Introduction}

There are numerous applications of automatic speech recognition (ASR)
in the present-day world. From personal mobile assistant apps for
the urban population to providing commodity price helplines for farmers
in rural areas, speech recognition has made its way from the research
laboratories to the common man. More important are the real-time interactive
voice response (IVR) based applications for information retrieval
and storage, such as automated primary health care, personalised farming,
price retrieval, ticket booking and so on.

Such applications, developed on low-end processors, require instant
speech-to-text conversion robustly under varying background noises
and microphone distortions. Numerous techniques have been proposed
to improve the noise robustness of speech recognition. These improved
accuracies come at the cost of higher computational complexity. The
use of low-end processors limit the processing capability, and also
consume time to perform intense computational tasks. This delays the
response of the system to the user, making it annoying and often unusable.
Thus only a few set of the techniques are suitable for robust real-time
ASR systems, and there is an interest in understanding and improving
them.

Typical back-end of an ASR system consists of a set of generative
hidden Markov models (HMMs) which are trained for each sound unit.
Features extracted from the user's speech data are compared against
the HMMs using a process called Viterbi decoding to get the most likely
word sequence that could have been spoken. To overcome the effect
of noise and distortion, a class of techniques \emph{normalise} the
features of test speech so that they have similar characteristics
to those of training.

\begin{singlespace}
This thesis concentrates on these feature normalisation techniques
suitable for improving robustness in real-time applications. Work
has been done on two scenarios, during the training process of the
back-end of the system, viz.,\vspace{-2.5ex}

\end{singlespace}
\begin{enumerate}
\begin{singlespace}
\item there is no information about noise
\item some noisy training speech data are available\end{singlespace}

\end{enumerate}

\section{Motivation}

Examples of feature normalisation techniques are cepstral mean and
variance normalisation (CMVN), histogram equalisation (HEQ), stereo-based
piecewise linear compensation for environments (SPLICE) etc. While
some of the techniques operate at utterance level and estimate statistical
parameters out of the test utterances, some operate on individual
feature vectors. Parameter estimation in the former set of techniques
requires sufficiently long utterances for reliable estimation, and
also some processing. The latter are best suited for shorter utterances
and real-time applications.

The standard features used for speech feature extraction are the Mel-frequency
cepstral coefficients (MFCCs). In the final stage of their extraction
process, a discrete cosine transform (DCT)\footnote{DCT, by default hereafter, refers to Type-II DCT}
is used for dimensionality reduction and feature decorrelation. An
interesting question to ponder at this juncture is: whether DCT is
the best method to achieve feature dimentionality reduction and feature
decorrelation along with noise robustness.

Heteroscedastic linear discriminant analysis (HLDA) \cite{nagendrathesis}
is one technique which achieves these qualities under low noise conditions,
even though it was originally designed for a different objective.
The technique fails in high noise environments. This thesis proposes
using non-negative matrix factorisation (NMF) \cite{leeseungnature}
and reconstruction of the features prior to applying the DCT in the
standard MFCC extraction process. Two methods of achieving noise robust
features are discussed in detail. These techniques do not assume any
information about noise during the training process.

It is also reasonable to identify the commonly encountered noise environments
during training, and learn their respective compensations. During
the real-time operation of the system, the background noise can be
classified into one of the known kinds of noise so that the corresponding
compensation can be applied. An example of one such technique is the
stereo-based piecewise compensation for environments (SPLICE) \cite{lideng}.
This technique requires stereo data during training, which consists
of simultaneously recorded speech using two microphones. One is a
close-talk microphone capturing mostly clean speech and the other
is a far-field microphone, which captures noise along with speech.

This technique is of particular interest because it operates on individual
feature vectors, and the compensation is an easily implementable linear
transformation of the feature. However, there are two disadvantages
of SPLICE. The algorithm fails when the test noise condition is not
seen during training. Also, owing to its requirement of stereo data
for training, the usage of the technique is quite restricted. This
thesis proposes a modified version of SPLICE that improves its performance
in all noise conditions, predominantly during severe noise mismatch.
An extension of this modification is also proposed for datasets that
are not stereo recorded, with minimal performance degradation as compared
to the conventional SPLICE. To further boost the performance, run-time
adaptation of the parameters is proposed, which is computationally
very efficient when compared to maximum likelihood linear regression
(MLLR), a standard model adaptation method.

\section{Overview of the Thesis}

The rest of the thesis is organised as follows. Chapter \ref{chap:Background}
summarises the required background and revises some existing techniques
in literature. Chapter \ref{chap:Non-Negative-Subspace-Projection}
discusses the proposed methods of performing feature compensation
using NMF during MFCC extraction, and assumes no information about
noise during training. Chapter \ref{chap:SPLICE} details the proposed
modifications and techniques using SPLICE. Finally, Chapter \ref{chap:Conclusion-and-Future}
concludes the thesis, indicating possible future extensions.

\chapter{Background\label{chap:Background}}

\section{HMM-GMM based Speech Recognition\label{sec:HMMs-for-Speech}}

The aim of a speech recognition system is to efficiently convert a
speech signal into a text transcription of spoken words \cite{rabinerspeechproc}.
This requires extracting relevant features from the largely available
speech samples, which is called feature extraction. By this process,
the speech signal is converted into a stream of feature vectors capturing
its time-varying spectral behaviour.

The feature vector streams of each basic sound unit are statistically
modelled as HMMs using a training process called Baum-Welch algorithm.
This requires sufficient amount of transcribed training speech. A
dictionary specifying the conversion of words into the basic sound
units is necessary in this process. During testing, an identical feature
extraction process, followed by Viterbi decoding are performed to
obtain the text output. The decoding is performed on a recognition
network built using the HMMs, dictionary and a language model. Language
model gives probabilities to sequences of words, and helps in boosting
the performance of the ASR system.

\begin{figure}
\begin{centering}
\includegraphics[scale=0.5]{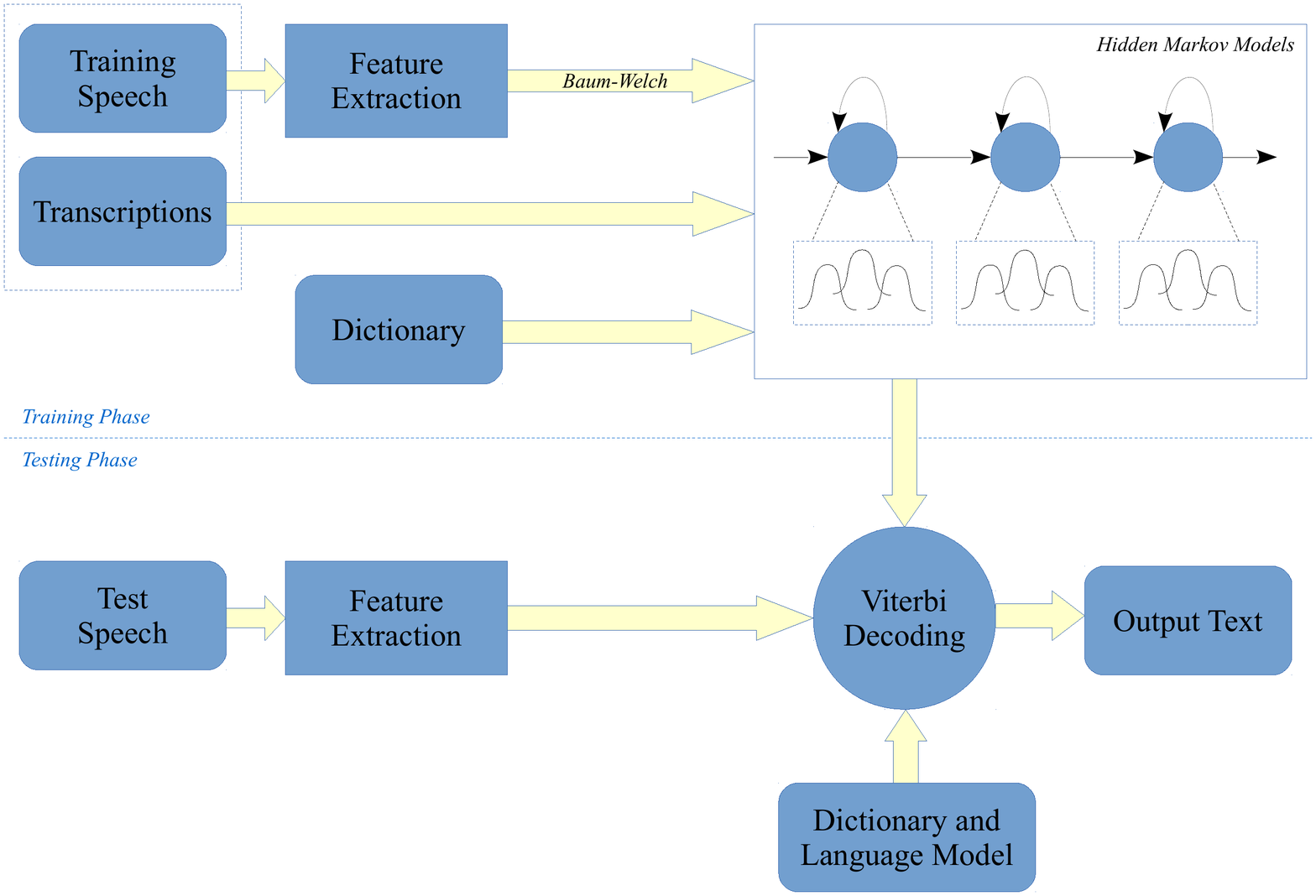}
\par\end{centering}

\protect\caption{Overview of speech recognition system\label{fig:ASR-Overview}}
\end{figure}

This process is summarised in Figure \ref{fig:ASR-Overview}. More
detailed explanation can be found in \cite{rabinerspeechproc}. The
choice of the basic sound unit depends on the size of the vocabulary.
Word models are convenient to use for tasks such as digit recognition.
For large vocabulary continuous ASR, triphone models are built, which
can be concatenated appropriately during Viterbi decoding to represent
words.

\section{MFCC Feature Extraction}

Apart from the information about what has been spoken, a speech signal
also contains distinct characteristics which vary with the recording
conditions such as background noise, microphone distortion, reverberation
and so on. Speaker dependent parameters such as vocal tract structure,
accent, mood and style of speaking also affect the signal. Thus an
ASR system requires robust feature extraction process which captures
only the speech information, discarding the rest.

\subsection{MFCCs}

MFCCs have been shown to be one of the effective features for ASR.
The extraction of MFCCs is summarised in Figure \ref{fig:MFCC-Extraction},
and involves the following steps:
\begin{enumerate}
\item \textbf{Short-time processing:} Convert the speech signal into overlapping
frames. On each frame, apply pre-emphasis to give weightage to higher
formants, apply a hamming window to minimise the signal truncation
effects, and take magnitude of the DFT.
\item \textbf{Log Mel filterbank (LMFB):} Apply a triangular filterbank
with Mel-warping and obtain a single coefficient output for each filter.
Apply log operation on each coefficient to reduce the dynamic range.
These operations are motivated by the acoustics of human hearing.
\item \textbf{DCT:} Apply DCT on each frame of LMFB coefficients. This provides
energy compaction useful for dimensionality reduction, as well as
approximate decorrelation useful for diagonal covariance modelling.
\begin{figure}[H]
\begin{centering}
\includegraphics[bb=0bp 0bp 992bp 531bp,clip,scale=0.45]{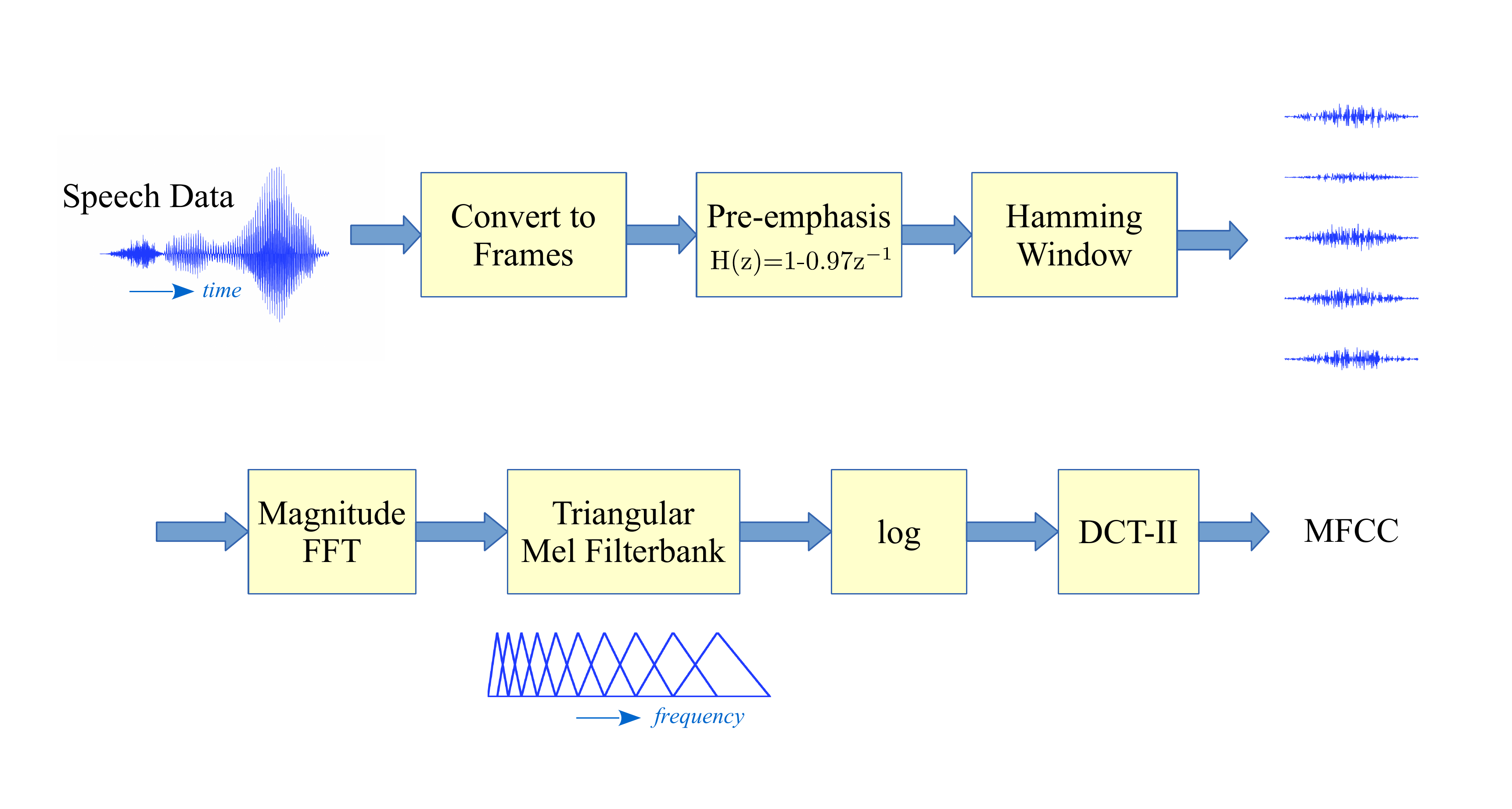}
\par\end{centering}

\protect\caption{Extraction process of MFCCs\label{fig:MFCC-Extraction}}
\end{figure}

\end{enumerate}
A cepstral lifter is generally used to give approximate equal weightage
to all the coefficients in MFCCs (the amplitude diminishes due to
energy compaction property of DCT). Delta and acceleration coefficients
are also appended to the liftered MFCCs, to capture the dynamic information
in the speech signal. Finally cepstral mean subtraction (CMS) is performed
on these composite features, to remove any stationary mismatch effects
caused by recording in different environments.

\section{Need for Additional Processing}

Let us look at the robustness aspects of MFCC composite features in
response to two kinds of undesired variations, viz., speaker-based
and environment-based.

During MFCC extraction, most of the pitch information is discarded
by the smoothing operation of LMFB. Some speaker-specific characteristics
are removed by truncating the higher cepstral coefficients after DCT
operation. However, other variations such as those occurring due to
differences in vocal-tract structures can be compensated for better
recognition results.

MFCC composite features are less robust to environment changes. The
presence of background noise, especially during testing, causes serious
non-linear distortions which cannot be compensated using CMS. This
\emph{acoustic mismatch} between the training and testing environments
needs additional compensation.

\subsection{Feature Compensation and Model Adaptation}

The techniques which operate on features to nullify the undesired
effects are called \emph{feature compensation} techniques. Examples
are CMVN, HEQ. These methods are usually simple and can be incorporated
into real-time applications. So there is an interest in understanding
and improving these techniques.

On the contrary, there are a class of \emph{model adaptation} techniques
which refine the models to compensate the effects. These are usually
computationally intense and yield high recognition rates. Examples
are maximum likelihood linear regression (MLLR), speaker adaptive
training (SAT).

Some techniques such as joint factor analysis (JFA), joint uncertainty
decoding (JUD) use a combination of both feature and model compensation
to further improve the recognition.

However, this thesis focuses on feature compensation done on frame-by-frame
basis, due to their suitability to real-time applications.

\section{A Brief Review of Some Techniques Used}

\subsection{HLDA}

Given features belonging to various \emph{classes}, this technique
aims at achieving discrimination among the classes through dimensionality
reduction. It linearly transforms $D$ dimensional features such that
\emph{only} $R\;\left(<D\right)$ dimensions of the transformed features
have the discriminating capability, and the remaining $D-R$ dimensions
can be discarded. In ASR, the features assigned to each state of an
HMM during training are typically considered as belonging to a class.
The transformation $\boldsymbol{\mathbf{\mathbf{\mathcal{H}}}}$ is
estimated from the training data using class labels obtained from
their first-pass transcriptions. The $R$ dimensional new features
are then used to train and test the models.

\noindent \textbf{Estimation of HLDA transform}

Let the feature $\mathbf{y}$ be transformed to obtain $\mathbf{x}=\boldsymbol{\mathbf{\mathbf{\mathcal{H}}}}\mathbf{y}$.
Let $\mathbf{y}^{(k)}$ denote that the class-label of $\mathbf{y}$
is $k$. For each class $k$, the set $\left\{ \mathbf{y}^{(k)}\right\} $
are assumed to be Gaussian, and thus are their corresponding $\left\{ \mathbf{x}^{(k)}\right\} $.
It is desired that the last $D-R$ dimensions of $\mathbf{x}$ do
not contain any discriminative information, i.e., the mean $\boldsymbol{\mu}_{k}$
and covariance $\boldsymbol{\Sigma}_{k}$ of all the $K$ classes
are identical in their last $D-R$ dimensions, as 
\[
\boldsymbol{\mu}_{k}=\begin{bmatrix}\widetilde{\boldsymbol{\mu}}_{k}\\
\boldsymbol{\mu}_{0}
\end{bmatrix},\quad\boldsymbol{\Sigma}_{k}=\begin{bmatrix}\widetilde{\boldsymbol{\Sigma}}_{k} & \mathbf{0}\\
\mathbf{0} & \boldsymbol{\Sigma}_{0}
\end{bmatrix}
\]
where $\boldsymbol{\mu}_{0}$ and $\boldsymbol{\Sigma}_{0}$ are of
dimensions $\left(D-R\right)\times1$ and $\left(D-R\right)\times\left(D-R\right)$
respectively. If this is achieved, the last $D-R$ dimensions of $\mathbf{x}^{(k)}$
can be discarded without loss of discriminability.

The density of each $\mathbf{y}^{(k)}$ is modelled as
\[
p(\mathbf{y}^{(k)})=\frac{|\boldsymbol{\mathbf{\mathcal{H}}}|}{(2\pi)^{\frac{D}{2}}|\boldsymbol{\Sigma}_{k}|^{\frac{1}{2}}}e^{-\frac{1}{2}(\mathbf{x}^{(k)}-\boldsymbol{\mu}_{k})^{T}\boldsymbol{\Sigma}_{k}^{-1}(\mathbf{x}^{(k)}-\boldsymbol{\mu}_{k})}
\]

Numerical solution for $\boldsymbol{\mathbf{\mathcal{H}}}$ can be
derived by differentiating the log-likelihood function $\mathcal{L}=\sum_{k=1}^{K}\log p(\mathbf{y}^{(k)})$
w.r.t $\boldsymbol{\mathbf{\mathcal{H}}}$, and using the maximum
likelihood estimates of $\boldsymbol{\mu}_{k}$ and $\boldsymbol{\Sigma}_{k}$
obtained from $\mathcal{L}$ \cite{nagendrathesis}.

\subsection{HEQ}

\begin{figure}[H]
\begin{centering}
\includegraphics[bb=0bp 0bp 539bp 482bp,clip,scale=0.5]{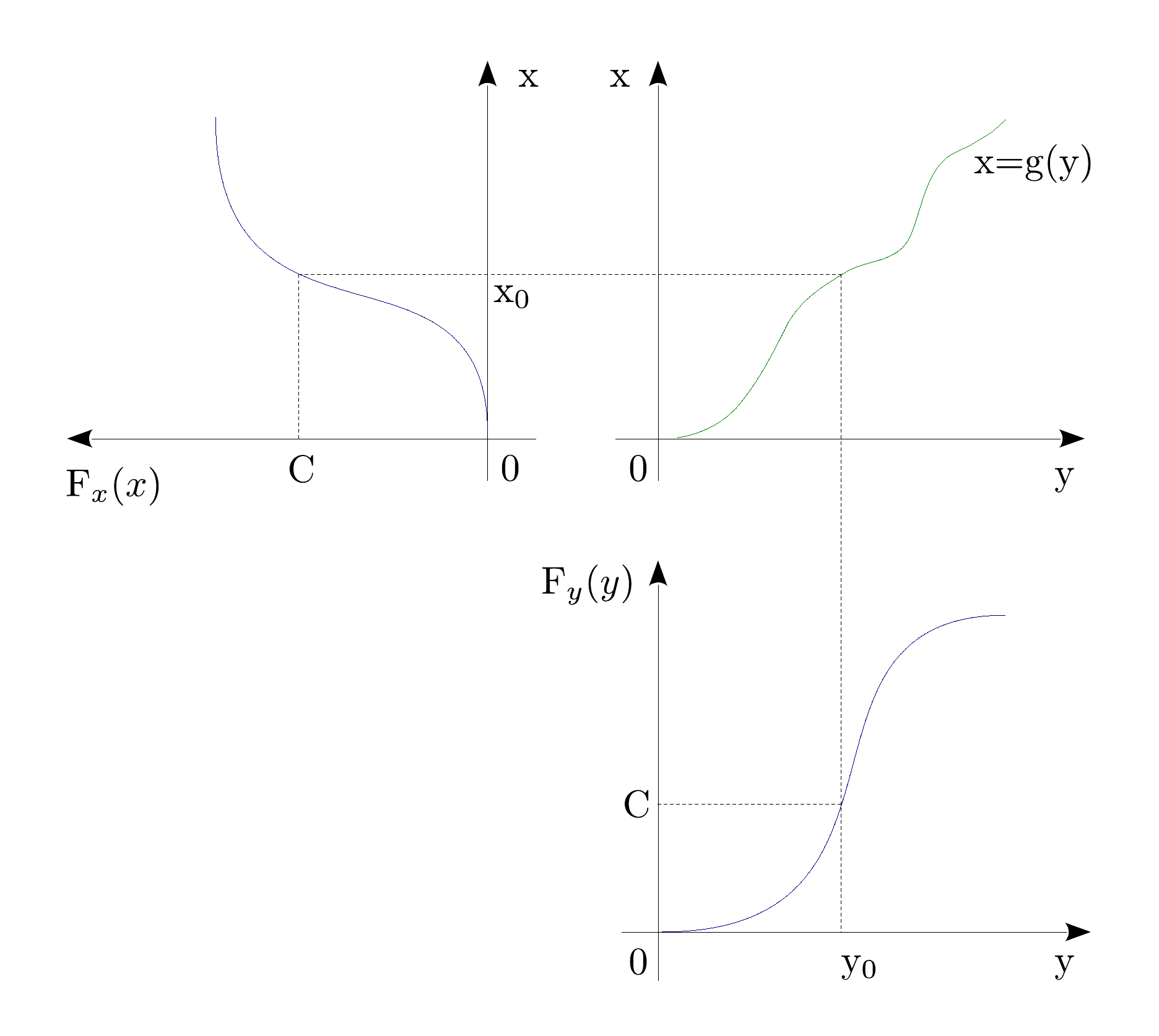}
\par\end{centering}

\protect\caption{Illustration of histogram equalisation\label{fig:Histogram-equalisation}}
\end{figure}

HEQ techniques are used to compensate for the acoustic mismatch between
the training and testing conditions of an ASR system, thereby giving
improved performance. HEQ is a feature compensation technique which
defines a transformation that maps the distribution of test speech
features onto a reference distribution. As shown in Figure \ref{fig:Histogram-equalisation},
a function $x=g\left(y\right)$ can be learned such that any noise
feature component $y_{0}$ can be transformed to its corresponding
equalised version $x_{0}$.

The reference distribution can be that of either clean speech, training
data or even a parametric function. Both the training and test features
need to be equalised to avoid mismatch. Since HEQ matches the whole
distribution, it matches the means, variances and all other higher
moments.

Let $y$ be a component of a noisy speech feature vector modelled
as $f_{Test}\left(y\right)$. The relation between the cumulative
distribution of $y$ and that of its equalised version $x$ is given
by
\[
\begin{aligned}F_{Test}\left(y\right) & =\int_{-\infty}^{y}f_{Test}\left(y'\right)dy'\\
 & =\int_{-\infty}^{g\left(y\right)}f_{Test}\left(g^{-1}\left(x'\right)\right)\frac{dg^{-1}\left(x'\right)}{dx'}dx'\\
 & =\int_{-\infty}^{x}f_{Train}\left(x'\right)dx'\;\bigg|_{x=g\left(y\right)}\\
 & =F_{Train}\left(x\right)
\end{aligned}
\]

In practice, the function $g\left(y\right)$ is implemented using
quantile-based methods \cite{quantileheqhilgerney}.

\subsection{MLLR}

MLLR is a widely used adaptation technique based on maximum-likelihood
principle \cite{legetterthesis,mllrgales}. It performs adaptation
through regression-based transformations on (usually) the mean vectors
$\boldsymbol{\mu}_{m}$ of the system of HMMs, $m$ being the mixture
index. The transformations are estimated such that the original system
is tuned to a new speaker or environment.
\[
\widetilde{\boldsymbol{\mu}}_{m}=\mathbf{\boldsymbol{\mathcal{M}}}_{m}\boldsymbol{\mu}'_{m}
\]
where $\mathbf{\boldsymbol{\mathcal{M}}}_{m}$ is an MLLR transformation
matrix of dimension $D\times\left(D+1\right)$, and $\boldsymbol{\mu}'_{m}=\begin{bmatrix}1 & \boldsymbol{\mu}_{m}^{T}\end{bmatrix}^{T}$.
An estimate of $\boldsymbol{\mathbf{\mathcal{M}}}_{m}$ is obtained
by maximising the likelihood function
\[
\mathbf{\mathcal{L}}=\frac{1}{(2\pi)^{\frac{D}{2}}|\boldsymbol{\Sigma}_{m}|^{\frac{1}{2}}}e^{-\frac{1}{2}(\mathbf{x}-\mathbf{\mathbf{\boldsymbol{\mathcal{M}}}_{m}}\boldsymbol{\mu}'_{m})^{T}\boldsymbol{\Sigma}_{m}^{-1}(\mathbf{x}-\mathbf{\boldsymbol{\mathcal{M}}}_{m}\boldsymbol{\mu}'_{m})}
\]

In this work, MLLR mean adaptation has been used as a \emph{global}
transform to adjust the system of HMMs to a new noise environment
encountered during testing. The adaptation data are the same as test
data, which consist of files recorded in a particular noise condition
with sufficient speaker and speech variations.

\subsection{SPLICE}

\begin{figure}
\begin{centering}
\includegraphics[clip,scale=0.5]{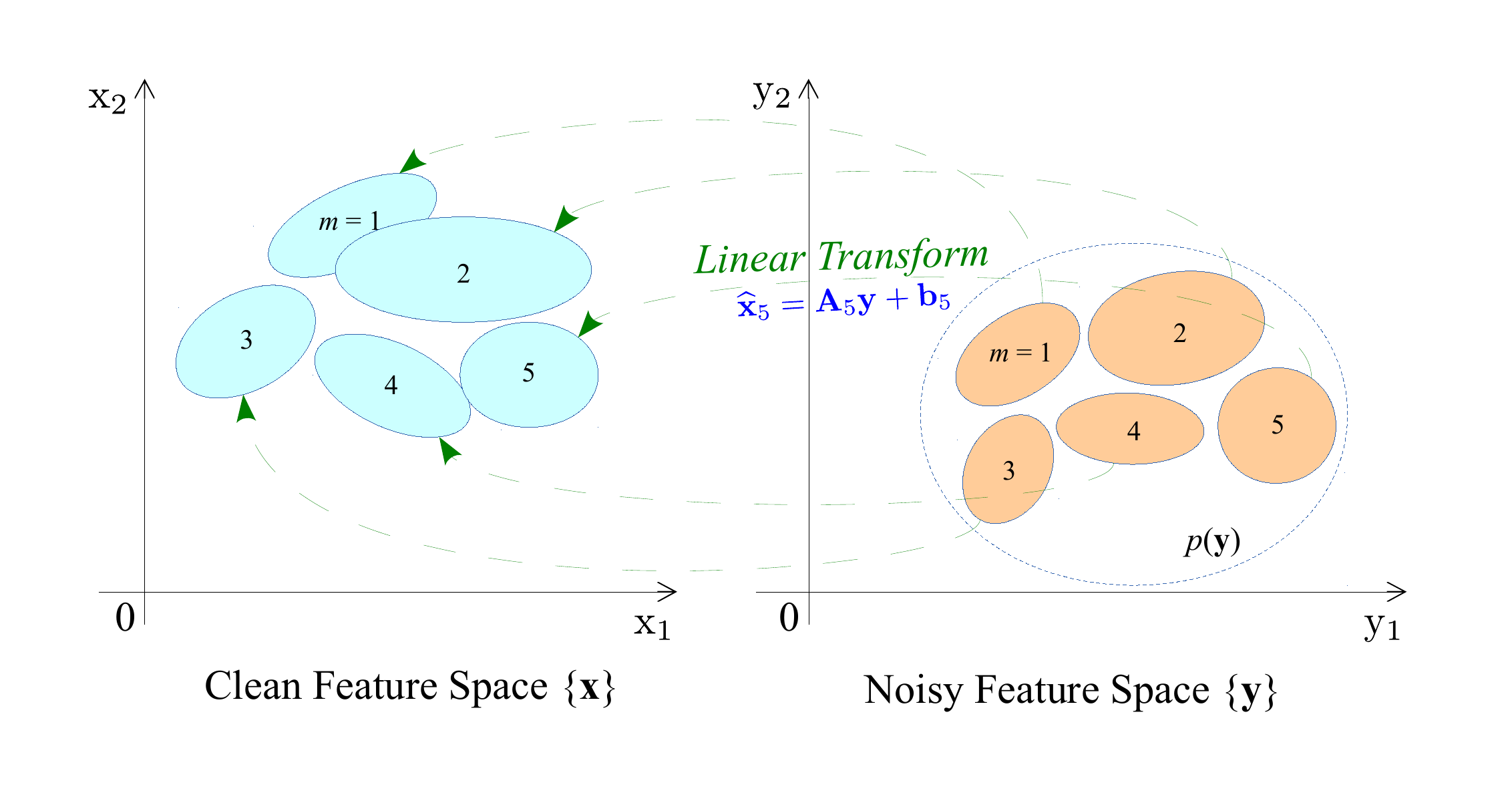}
\par\end{centering}

\protect\caption{Illustration of SPLICE feature enhancement\label{fig:SPLICE-feature-enhancement}}
\end{figure}

SPLICE is a popular and efficient noise robust feature enhancement
technique. It partitions the noisy feature space into $M$ classes,
and learns a linear transformation based noise compensation for each
partition class during training, using stereo data. Any test vector
$\mathbf{y}$ is soft-assigned to one or more classes by computing
$p\left(m\,|\,\mathbf{y}\right)$ $\left(m=1,2,\ldots,M\right)$,
and is compensated by applying the weighted combination of linear
transformations to get the \emph{cleaned} version $\widehat{\mathbf{x}}$.
\begin{equation}
\widehat{\mathbf{x}}=\sum_{m=1}^{M}p\left(m\,|\,\mathbf{y}\right)\left(\mathbf{A}_{m}\mathbf{y}+\mathbf{b}_{m}\right)\label{eq:splice}
\end{equation}
$\mathbf{A}_{m}$ and $\mathbf{b}_{m}$ are estimated during training
using stereo data. The training noisy vectors $\{\mathbf{y}\}$ are
modelled using a Gaussian mixture model (GMM) $p(\mathbf{y})$ of
$M$ mixtures, and $p\left(m\,|\,\mathbf{y}\right)$ is calculated
for a test vector as a set of posterior probabilities w.r.t the GMM
$p(\mathbf{y})$. Thus the partition class is decided by the mixture
assignments $p\left(m\,|\,\mathbf{y}\right)$. This is illustrated
in Figure \ref{fig:SPLICE-feature-enhancement}.

\subsection{NMF}

NMF is an approximate matrix decomposition 
\begin{equation}
\underset{\left(D\times N\right)}{\mathbf{V}}\approx\underset{\left(D\times R\right)}{\mathbf{W}}\underset{\left(R\times N\right)}{\mathbf{H}}\label{eq:nmf}
\end{equation}
where $\mathbf{V}$, consisting of non-negative elements, is decomposed
into two non-negative matrices $\mathbf{W}$ and $\mathbf{H}$. In
the context of speech data, the columns of $\mathbf{V}$ constitute
non-negative feature vectors $\{\mathbf{v}_{n}\}\in\mathbb{R}^{D+},\,n=1,2,\ldots,N$.
 After the decomposition, $\mathbf{W}$ is a non-negative dictionary
of basis vectors $\{\mathbf{w}_{r}\}\in\mathbb{R}^{D+},\,r=1,2,\ldots,R$
along columns, representing a useful subspace in which $\{\mathbf{v}_{n}\}$
are contained, when $R<D$. $\mathbf{H}\in\mathbb{R}^{(R\times N)+}$
is a matrix of vectors $\{\mathbf{h}_{n}\}\in\mathbb{R}^{R+},n=1,2,\ldots,N$
such that each $\mathbf{h}_{n}=\{h_{1n},h_{2n},\ldots,h_{Rn}\}$ is
a set of weights or \emph{activation coefficients} acting upon all
the bases $\{\mathbf{w}_{r}\}$ to give the corresponding $\mathbf{v}_{n}$,
independently of all the other columns, as shown in equation \ref{eq:nmfvectorial}.
\begin{equation}
\mathbf{v}_{n}\approx\sum_{r=1}^{R}\mathbf{w}_{r}h_{rn}\label{eq:nmfvectorial}
\end{equation}

The decomposition (\ref{eq:nmf}) can be learned by randomly initialising
$\mathbf{W}$ and $\mathbf{H}$. The estimates of $\mathbf{W}$ and
$\mathbf{H}$ are iteratively improved by minimising a KL-divergence
based cost function

\begin{equation}
\mathcal{D}(\mathbf{V}\parallel\mathbf{WH})=\sum_{d,n}\left[\mathbf{V}\otimes\log\left(\frac{\mathbf{V}}{\mathbf{W}\mathbf{H}}\right)-\mathbf{V}+\mathbf{WH}\right]_{dn}\label{eq:kldcost}
\end{equation}
 where $\otimes$ denotes Hadamard product, the division inside the
log is element-wise and the summation is over all the elements of
the matrix. The optimisation 
\[
\underset{\mathbf{W},\mathbf{H}}{\arg\min}\enskip\mathcal{D}(\mathbf{V}\parallel\mathbf{WH})\qquad\mathbf{W}>0,\enskip\mathbf{H}>0
\]
 gives the following iterative update rules \cite{leeseungnature,leeseung}
for refining the matrices $\mathbf{W}$ and $\mathbf{H}$: 
\begin{equation}
w_{dr}:=w_{dr}\frac{\sum_{n}h_{rn}v_{dn}/[\mathbf{WH}]_{dn}}{\sum_{n}h_{rn}}\label{eq:Wupdate}
\end{equation}
 
\begin{equation}
h_{rn}:=h_{rn}\frac{\sum_{d}w_{dr}v_{dn}/[\mathbf{WH}]_{dn}}{\sum_{d}w_{dr}}\label{eq:Hupdate}
\end{equation}
 where ``$:=$'' refers to assignment operator. It can be seen that
the update rules are multiplicative, i.e., the matrices are updated
by performing just a product with another matrix, making the implementation
simple and quick. The columns of $\mathbf{W}$ can be thought of as
basic building blocks that can reconstruct speech features.

\section{Recent Work in Literature - Motivation}

\cite{exemplarasr} showed an overview of a wide range of techniques
in ASR that use speech \emph{exemplars} (dictionaries). It was argued
that noise robustness can be achieved through the use of speech dictionaries.
However, most of these techniques are computationally intense.

Since NMF is one of the methods of learning dictionaries and is easily
implementable, it is of particular interest. NMF is known to learn
useful time-frequency patterns in a given dataset, and has been applied
to learn spectral representations in audio and speech applications
that include audio source separation, supervised speech separation,
speech enhancement and recognition. A few of them are mentioned below.

A regularised variant of NMF was used by \cite{nmfpriors} to learn
separate dictionaries of speech and noise. The concatenated dictionary
was used in NMF to learn weights of noisy test utterances, where the
weights corresponding to noise bases are suppressed to achieve speech
denoising and thus enhancement. \cite{nnmfrobust} proposed supervised
NMF for improving noise robustness in spelling recognition task. NMF
was performed using a predetermined $\mathbf{W}$ consisting of spectra
of spelled letters. The authors showed that appending the weights
$\mathbf{H}$ to MFCC and BLSTM (Bidirectional Long Short-Term Memory
neural network) features improves noise robustness. \cite{gemmeke}
represented LMFB features of noisy speech using exemplars of speech
and noise bases. A hybrid (exemplar and HMM based) recognition on
Aurora-2 task was performed to achieve noise robustness at high noise
levels.

Most of the above techniques have used the weights $\mathbf{H}$ as
new features or combining them with existing features for improved
recognition. In contrast, Chapter \ref{chap:Non-Negative-Subspace-Projection}
of this thesis will show that multiplying back $\mathbf{W}$ and $\mathbf{H}$
to get new LMFB features and converting them to MFCCs improves noise
robustness. This approach is useful in real-time applications because
of its fast implementation. A technique which learns a robust $\mathbf{W}$
will also be discussed. These methods do not assume any information
about noise during training.

When there are noisy training files available, techniques such as
SPLICE learn compensation from the seen noisy data to obtain their
corresponding clean versions. Over the last decade, improvements using
uncertainty decoding \cite{udsplice}, maximum mutual information
based training \cite{mmisplice}, speaker normalisation \cite{spkrnormsplice}
etc. were introduced in SPLICE framework. There are two disadvantages
of SPLICE. The algorithm fails when the test noise condition is not
seen during training. Also, owing to its requirement of stereo data
for training, the usage of the technique is quite restricted. So there
is an interest in addressing these issues.

\cite{unseensplice} recently proposed an adaptation framework using
Eigen-SPLICE to address the problem of unseen noise conditions. The
method involves preparation of quasi stereo data using the noise frames
extracted from non-speech portions of the test utterances. For this,
the recognition system is required to have access to some clean training
utterances for performing run-time adaptation.

\cite{gonzalez} proposed a stereo-based feature compensation method,
which is similar to SPLICE in certain aspects. Clean and noisy feature
spaces were partitioned into vector quantised (VQ) regions. The stereo
vector pairs belonging to $i^{th}$ VQ region in clean space and $j^{th}$
VQ region in noisy space are classified to the $ij^{th}$ sub-region.
Transformations based on Gaussian whitening expression were estimated
from every noisy sub-region to clean sub-region. But it is not always
guaranteed to have enough data to estimate a full transformation matrix
from each sub-region to other.

In Chapter \ref{chap:SPLICE}, a simple modification to SPLICE will
be proposed, based on an assumption made on the correlation of training
stereo data. This will be shown to give improved performance in all
the noise conditions, predominantly in unseen conditions which are
highly mismatched with those of training. An extension of the method
to non-stereo datasets (which are not stereo recorded) will be proposed,
with minimal performance degradation as compared to conventional SPLICE.
Finally, an MLLR-based run-time noise adaptation framework will be
proposed, which is computationally efficient and achieves better results
than MLLR-based model adaptation.

\chapter{Non-Negative Subspace Projection\label{chap:Non-Negative-Subspace-Projection}}

\section{Introduction}

During MFCC extraction, usually 23 LMFB coefficients are converted
into 13 dimensional MFCCs through DCT. Results in Table \ref{tab:NMFresults}
show that performing HLDA on 39 dimensional MFCCs gives better robustness
to noise than MFCCs when the noise levels are low. But this technique
fails in high noise conditions. However, the objective of HLDA is
not to achieve robustness, but class-separability. Here a method is
proposed which aims at finding representations of speech feature vectors
using building-blocks.
\begin{figure}
\begin{centering}
\includegraphics[bb=0bp 0bp 397bp 283bp,clip,scale=0.6]{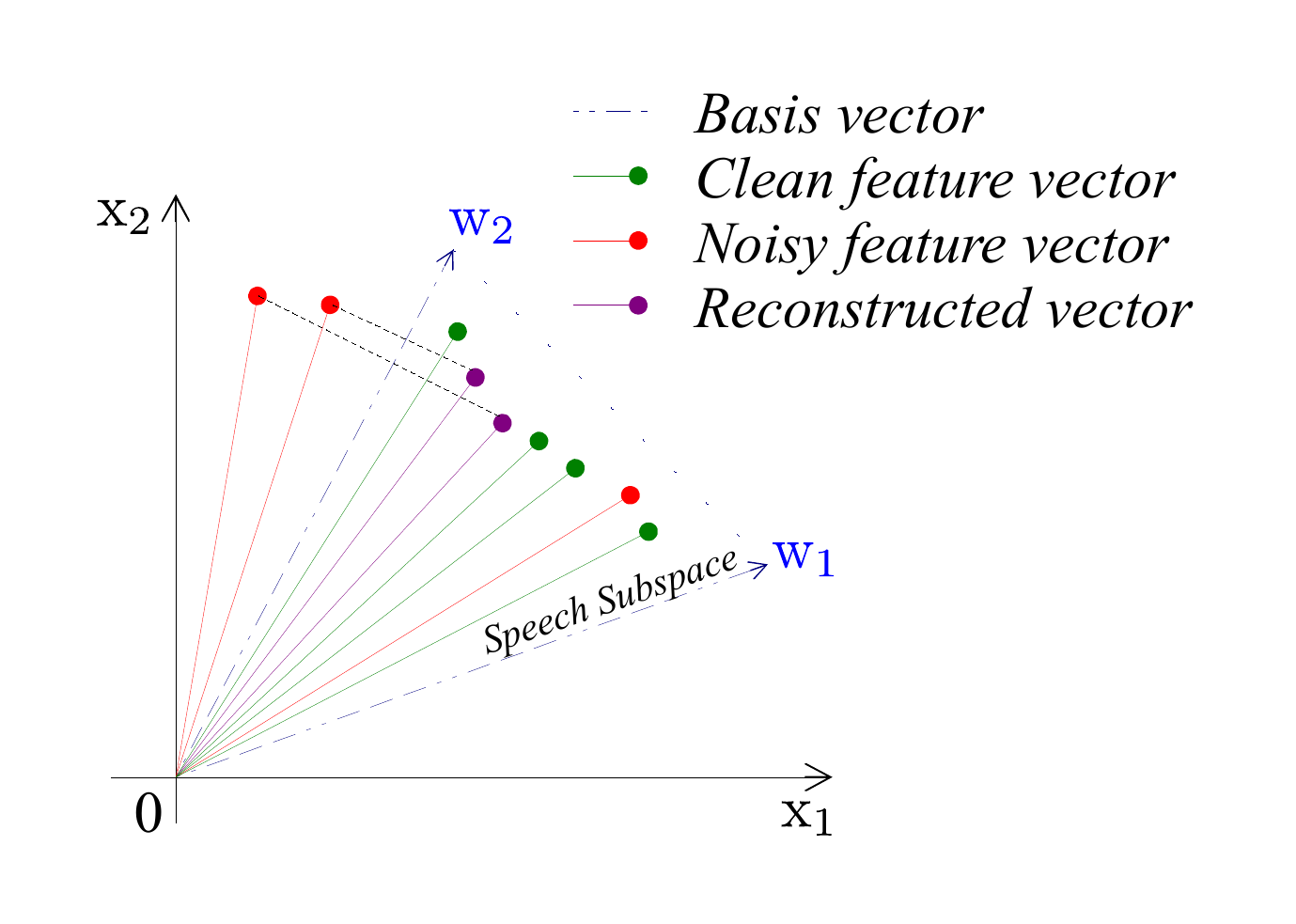}
\par\end{centering}

\protect\caption{NMF subspace projection\label{fig:NMF-subspace-projection}}
\end{figure}

The method operates on LMFB feature vectors by representing them using
non-negative linear combinations of their non-negative building blocks.
DCT-II is applied on the new feature vectors to obtain new MFCCs.
This incorporates, into the MFCC extraction process, a concept that
the speech features are made up of underlying building blocks. Apart
from the proposed additional step, conventional MFCC extraction framework
is maintained throughout the process. The building blocks of speech
are learned using NMF on speech feature vectors. Experimental results
show that the new MFCC features are more robust to noise, and achieve
better results when combined with the existing noise robust feature
normalisation techniques like HEQ and HLDA.

\section{The Speech Subspace}

The columns of $\mathbf{W}$ can be thought of as basic building blocks
that construct all the speech feature vectors, or the bases for subspace
of the speech feature vectors. So far, no mathematical proof has been
derived to validate if NMF learns the representations of the underlying
data. However, much of the literature support this. \cite{smaragdisconv}
states that the basis functions describe the spectral characteristics
of the input signal components, and reveal its \emph{vertical structure}.
\cite{nmfpriors} refer to these building blocks as a set of spectral
shapes. \cite{nnmfrobust} refer to them as spectra of events occurring
in the signal, and NMF is known to learn useful time-frequency patterns
in a given dataset.

One could probably use other basis learning techniques like principal
component analysis (PCA) to estimate the clean subspace. Though PCA
finds the directions of maximum spread of the features, these directions
need not contain only speech information. In methods such as NMF,
new features can be reconstructed within the subspace using different
cost functions to achieve desired qualities such as source localisation,
sparseness etc. KL-divergence based NMF has been successfully applied
in speech applications \cite{nmfpriors,nnmfrobust} over the conventional
Euclidean distance measure. In addition, the non-negativity constraint
is an added advantage. Figure \ref{fig:NMF-subspace-projection} shows
the bases $\mathbf{w}_{1}$ and $\mathbf{w}_{2}$ learned by NMF on
two-dimensional data contained in the subspace shown. Any vector outside
this subspace cannot be reconstructed perfectly using $\mathbf{w}_{1}$
and $\mathbf{w}_{2}$ when the weights acting on them are constrained
to be non-negative. So when these features are moved away from the
subspace due to the effect of noise, the reconstructed features can
be used in place of these features as better representations of the
underlying signal. However, vectors still inside the subspace are
not compensated.
\begin{figure}
\begin{centering}
\subfloat[Original features\label{fig:speechEg}]{\protect\begin{centering}
\protect\includegraphics[scale=0.5]{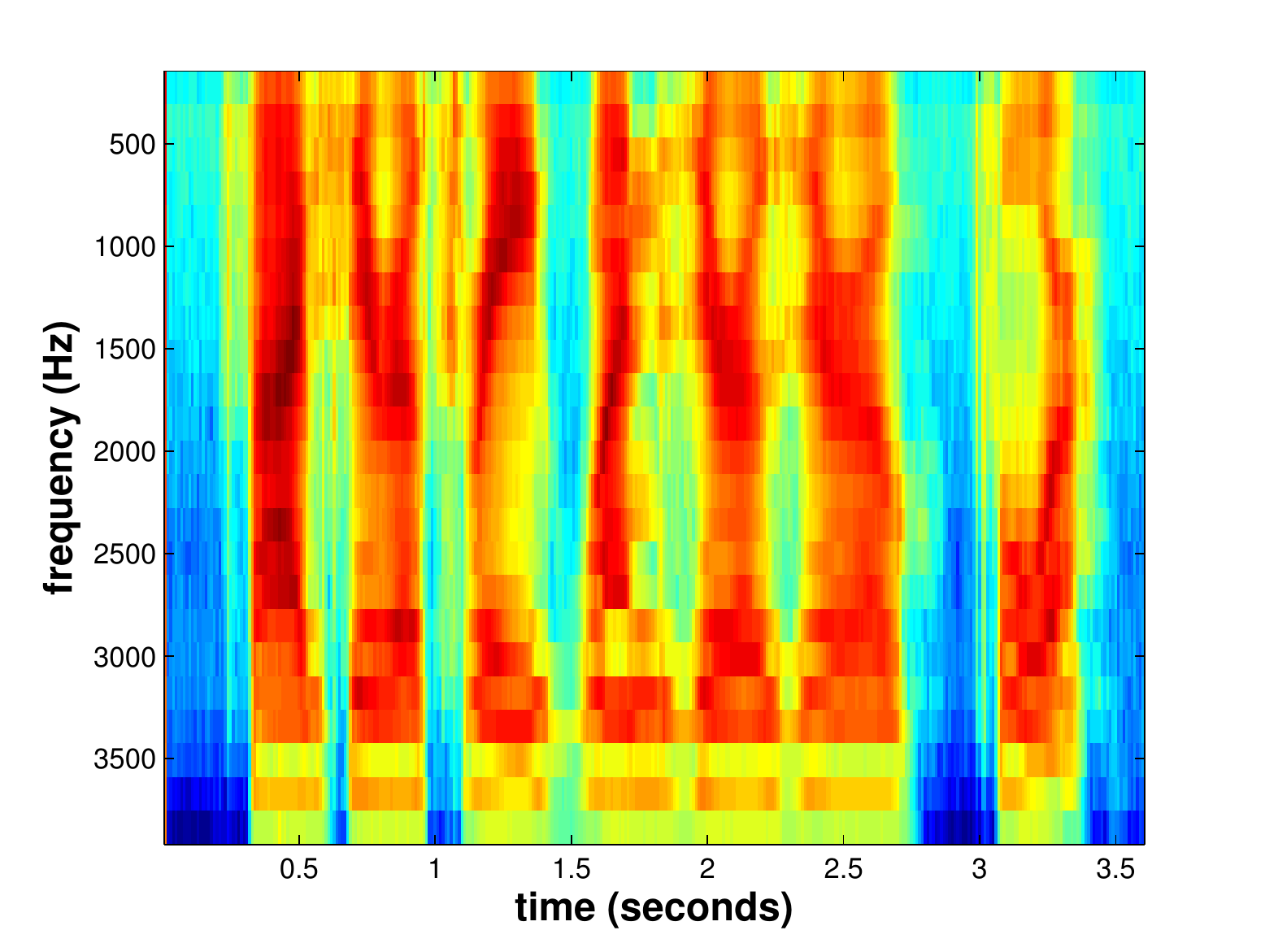}\protect
\par\end{centering}

}
\par\end{centering}

\begin{centering}
\subfloat[Reconstruction using NMF individual basis vectors\label{fig:EgReconstruction}]{\protect\begin{centering}
\protect\includegraphics[scale=0.5]{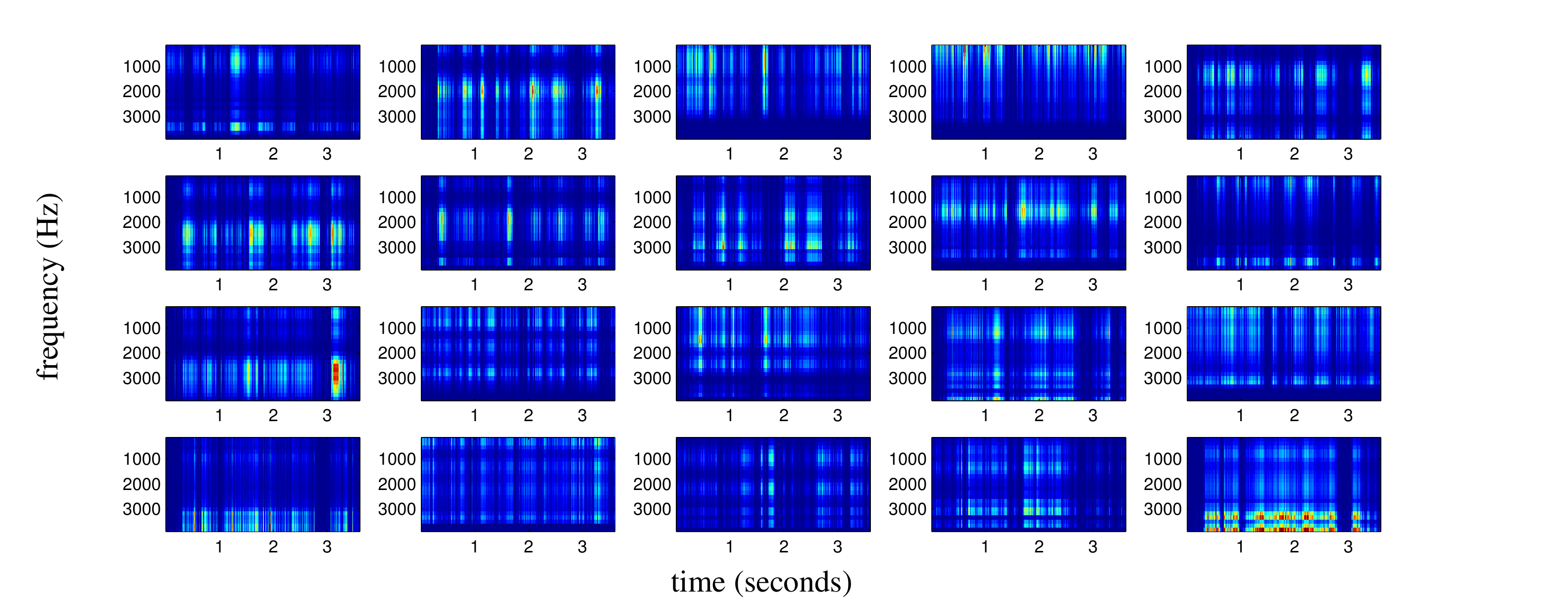}\protect
\par\end{centering}

}
\par\end{centering}

\protect\caption{LMFB features of a speech utterance}
\end{figure}

\section{Learning the Speech Subspace\label{sec:LearningWs}}

NMF can be performed by optimising cost functions based on different
measures such as Euclidean, KL-divergence, Itakura-Saito distances.
Here KL-divergence based method is chosen, which gives the update
equations (\ref{eq:Wupdate}) and (\ref{eq:Hupdate}). The decomposition
is not unique since the cost function (\ref{eq:kldcost}) to be optimised
is not convex. Depending on the application, one may choose to perform
update on both $\mathbf{W}$ and $\mathbf{H}$ in each iteration,
or fix one matrix and update the other. While simultaneously refining
$\mathbf{W}$ and $\mathbf{H}$, the columns of $\mathbf{W}$ may
be normalised after each update so that the sum of each column adds
up to value $1$. The scaling is automatically compensated in weight
matrix $\mathbf{H}$ during its update.

Fig. \ref{fig:speechEg} shows the plot of LMFB outputs of an utterance
containing connected spoken digits. Using the dictionary learned from
Aurora-2 database (as described in Section \ref{sub:NMF_plain}),
the reconstructions of the same utterance from each individual basis
vector of $\mathbf{W}$ are plotted in Fig. \ref{fig:EgReconstruction}.
It can be observed that each of the reconstructions has only a set
of particular dominant frequencies that are captured by the corresponding
basis vector. Such a dictionary can capture useful combinations of
frequencies present in speech utterances, and can give noise-robust
speech reconstructions. The reconstructed feature vectors are more
correlated, due to their confinement to the speech subspace, than
the original ones.

In PCA method, the speech dictionary $\mathbf{W}$ can be built by
stacking the most significant Eigen vectors of the clean training
data as columns. Any speech feature $\mathbf{v}$ can be reconstructed
as
\[
\widehat{\mathbf{v}}=\mathbf{W}\mathbf{W}^{T}\mathbf{v}
\]

\section{Proposed Feature Extraction Methods\label{sec:ProposedMethods}}

\begin{figure}
\begin{centering}
\includegraphics[scale=0.55]{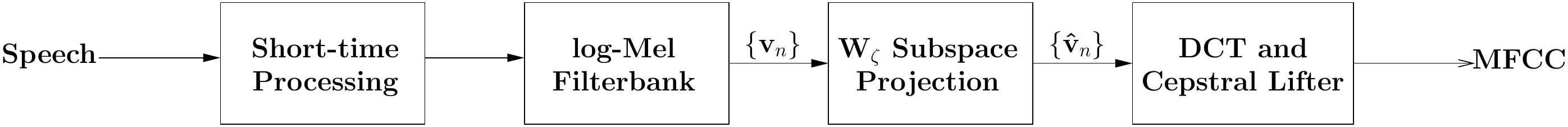}
\par\end{centering}

\protect\caption{Computation of subspace projected MFCCs\label{fig:nmfmfcc}}
\end{figure}

Speech signal is passed through conventional short-time processing
steps followed by an LMFB. Before applying DCT, which corresponds
to obtaining the conventional MFCC features, an additional step is
proposed to be introduced as shown in Fig. \ref{fig:nmfmfcc}, so
that the new MFCCs obtained after the processing are more robust to
noise. The additional step is computed in two different methods as
described in Sections \ref{sub:NMF_plain} and \ref{sub:NMF_robustW},
and the performances are compared in Sections \ref{sec:Experimental-Results}.

Short-time processing of speech signal includes applying STFT with
Hamming window of length 25$ms$ at a frame rate of 100 frames/second,
followed by passing the mean subtracted frames through a pre-emphasis
filter $1-0.97z^{-1}$. A conventional Mel-filter bank of $23$ constant
bandwidth equal gain linear triangular filters on the Mel scale is
applied to get a set of filter outputs for each frame. These outputs
are Mel-floored to value 1.0, and log operator is applied to obtain
non-negative LMFB feature vectors $\{\mathbf{v}_{n}\}$.

Let the LMFB feature vectors $\{\mathbf{v}_{\zeta}\}$ and $\{\mathbf{v}_{\varphi}\}$
corresponding to the training and test speech be stacked as columns
of $\mathbf{V}_{\zeta}$ and $\mathbf{V}_{\varphi}$ respectively.

\subsection{NMF\_plain\label{sub:NMF_plain}}

$\mathbf{V}_{\zeta}$ is decomposed as $\mathbf{V}_{\zeta}\approx\mathbf{W}_{\zeta}\mathbf{H}_{\zeta}$
using NMF decomposition by simultaneously updating $\mathbf{W}_{\zeta}$
and $\mathbf{H}_{\zeta}$ by (\ref{eq:Wupdate}) and (\ref{eq:Hupdate}).
Each column of $\mathbf{W}_{\zeta}$ is normalised after every iteration.
Finally, $\mathbf{W}_{\zeta}$ learns the building blocks of these
feature vectors. During testing, $\mathbf{V}_{\varphi}$ is approximated
as the product of $\mathbf{W}_{\zeta}$ and $\mathbf{H}_{\varphi}$
(i.e., $\mathbf{V}_{\varphi}\approx\mathbf{W}_{\zeta}\mathbf{H}_{\varphi}$).
This is done by fixing $\mathbf{W}_{\zeta}$ obtained during training
and performing update on $\mathbf{H}_{\varphi}$ alone using (\ref{eq:Hupdate}).
The new feature vectors in training and testing are thus 
\begin{equation}
\widehat{\mathbf{V}}_{\zeta}=\mathbf{W}_{\zeta}\mathbf{H}_{\zeta}\label{eq:vhattr}
\end{equation}
 and 
\begin{equation}
\mathbf{\widehat{V}}_{\varphi}=\mathbf{W}_{\zeta}\mathbf{H}_{\varphi}\label{eq:vhatte}
\end{equation}
 respectively. In the implementation, the whole training data has
been taken as $\mathbf{V}_{\zeta}$ to compute $\mathbf{W}_{\zeta}$
using NMF. However, during testing, each test utterance can separately
be taken as $\mathbf{V}_{\varphi}$ and the corresponding $\mathbf{H}_{\varphi}$
can be computed, fixing $\mathbf{W}_{\zeta}$.

Here the assumption is that many of the columns of $\mathbf{V}_{\varphi}$
are initially outside the speech subspace due to the effect of noise.
If each of these vectors outside the subspace is mapped to the \emph{nearest}
vectors $\widehat{\mathbf{V}}_{\varphi}$ within the subspace, the
new features are more noise-robust. If noise moves a clean feature
vector to another vector in the same subspace, it cannot be compensated.
Here the subspace is captured by $\mathbf{W}_{\zeta}$, and the term
\emph{nearest} is meant in the measure of KL-divergence \emph{distance}.
Replacement of $\mathbf{V}_{\varphi}$ by $\widehat{\mathbf{V}}_{\varphi}$
can alternatively be justified as follows. Each $\mathbf{v}_{\varphi}$
is being represented by non-negative linear combinations of building
blocks $\mathbf{w}_{\zeta}$ of speech data, because of which any
noise component in $\mathbf{v}_{\varphi}$ cannot be reconstructed
using speech bases, and hence cannot be retained in $\mathbf{\widehat{v}}_{\varphi}$.

Now, by convention, DCT matrix $\mathbf{D}$, followed by a standard
cepstral lifter $\mathbf{L}$ are applied on the features $\mathbf{V}_{\zeta}$
and $\mathbf{V}_{\varphi}$ to obtain 13 dimensional MFCC features
($C_{0}\ldots C_{12}$) for each frame, given by Eqs. (\ref{eq:ceptr})
and (\ref{eq:cepte}). 
\begin{equation}
\mathbf{C}_{\zeta}=\mathbf{LD}\widehat{\mathbf{V}}_{\zeta}\label{eq:ceptr}
\end{equation}
\begin{equation}
\mathbf{C}_{\varphi}=\mathbf{LD}\widehat{\mathbf{V}}_{\varphi}\label{eq:cepte}
\end{equation}

These features are used to parameterise the acoustic models (HMMs),
as explained in Section \ref{sub:NMF-Experimental-Setup}.

\subsection{NMF\_robustW\label{sub:NMF_robustW}}

\begin{figure}
\begin{centering}
\includegraphics[scale=0.55]{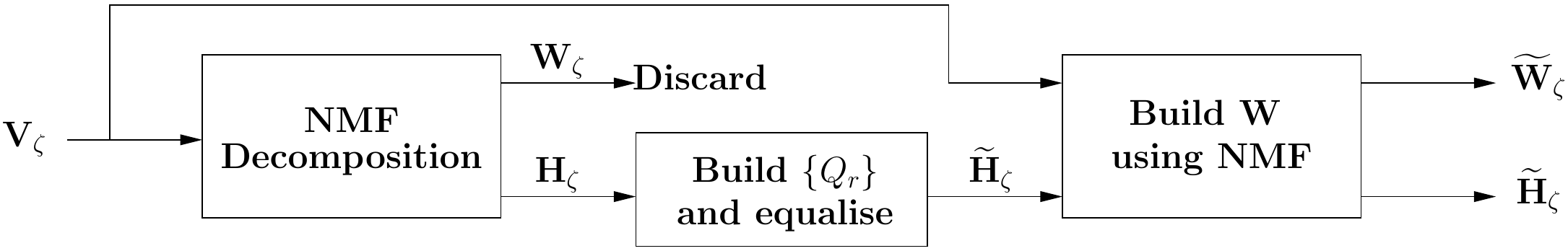}
\par\end{centering}

\protect\caption{Learning the speech dictionary in NMF\_robustW\label{fig:nmfbetterws}}
\end{figure}

Each set of basis activation coefficients $\mathbf{h}_{n}$ (for both
training and test data) is unique for the corresponding speech frame.
An addition of noise changes the statistics of $\mathbf{h}_{n}$.
So it is intuitive that equalising the statistics of $\mathbf{H}_{\varphi}$
during testing, to match that of training $\mathbf{H}_{\zeta}$, improves
the recognition. The equalisation has to be applied during training
also, to avoid mismatch of the test features against the built acoustic
model. Here the intention is not to perform test feature equalisation
explicitly, but to get a dictionary $\widetilde{\mathbf{W}}_{\zeta}$
which helps learn the test weights that are in equalised form, and
thus are more robust. Figure \ref{fig:nmfbetterws} shows a method
of obtaining the \emph{better} dictionary $\widetilde{\mathbf{W}}_{\zeta}$
from $\mathbf{W}_{\zeta}$, $\mathbf{H}_{\zeta}$ and $\mathbf{V}_{\zeta}$,
using HEQ during training.

After decomposing $\mathbf{V}_{\zeta}$ as $\mathbf{V}_{\zeta}\approx\mathbf{W}_{\zeta}\mathbf{H}_{\zeta}$
using NMF, a set of $R$ reference histograms $\left\{ Q_{r}\right\} $
of $\mathbf{H}_{\zeta}$ are calculated, one for each of its feature
component. HEQ is performed on $\mathbf{H}_{\zeta}$, as described
in Section \ref{sec:Cascading}, to get \emph{better} or \emph{more
robust} activations $\mathbf{\widetilde{H}}_{\zeta}$. But $\mathbf{W}_{\zeta}\mathbf{\widetilde{H}}_{\zeta}$
no longer matches $\mathbf{V}_{\zeta}$. So a \emph{better} $\mathbf{W}_{\zeta}$
is estimated using NMF update (\ref{eq:Wupdate}), i.e., by fixing
$\mathbf{\widetilde{H}}_{\zeta}$, $\mathbf{V}_{\zeta}$ and updating
$\mathbf{W}$. This essentially solves the optimisation 
\[
\widetilde{\mathbf{W}}_{\zeta}=\underset{\mathbf{W}}{\arg\min}\enskip D(\mathbf{V_{\zeta}}\parallel\mathbf{W\widetilde{H}_{\zeta}})\qquad\mathbf{W}>0,\enskip\mathbf{\widetilde{H}_{\zeta}}>0
\]

In other words, the speech dictionary is chosen such that the weights
become statistically equalised during training. Experimental results
show that $\widetilde{\mathbf{W}}_{\zeta}$ is a better dictionary
than $\mathbf{W}_{\zeta}$ in terms of noise-robustness. The training
features are thus 
\begin{equation}
\widehat{\mathbf{V}}_{\zeta}=\mathbf{\widetilde{W}}_{\zeta}\mathbf{\widetilde{H}}_{\zeta}\label{eq:vhattrnmf2}
\end{equation}

Testing process is the same as described in Section \ref{sub:NMF_plain},
except that the new $\widetilde{\mathbf{W}}_{\zeta}$ is directly
used to estimate the test data weights $\mathbf{H}_{\varphi}$, instead
of $\mathbf{W}_{\zeta}$. Additional equalisation is not performed
during testing, since the dictionary itself helps in learning equalised
weights. The test features are thus 
\begin{equation}
\mathbf{\widehat{V}}_{\varphi}=\mathbf{\widetilde{W}}_{\zeta}\mathbf{H}_{\varphi}\label{eq:vhattenmf2}
\end{equation}

As per convention, DCT and cepstral lifter matrices are applied on
$\widehat{\mathbf{V}}_{\zeta}$ and $\mathbf{\widehat{V}}_{\varphi}$,
given by Eqs. (\ref{eq:ceptr}) and (\ref{eq:cepte}) to obtain the
13 dimensional MFCC feature vectors for training and testing the HMMs.
Here it can be seen that the computational cost of both NMF\_plain
and NMF\_robustW are the same during testing.

\section{Cascading with Existing Techniques\label{sec:Cascading}}

The proposed methods can be cascaded with HEQ, where reference histogram
can be built using $\mathbf{C}_{\zeta}$, and HEQ can be applied on
$\mathbf{C}_{\zeta}$ and $\mathbf{C}_{\varphi}$ to get new features
vectors, using which acoustic models can be built as described in
Section \ref{sub:NMF-Experimental-Setup}.

Since the feature vectors given by Eqs. (\ref{eq:vhattr}) and (\ref{eq:vhatte})
are constrained to the column space of $\mathbf{W}_{\zeta}$, there
is certain additional correlation introduced in them. Their corresponding
cepstral features $\mathbf{C}_{\zeta}$ and $\mathbf{C}_{\varphi}$
are made approximately decorrelated through the use of DCT. However,
an additional decorrelation using HLDA is expected to further reduce
the feature correlations, besides utilising the advantage of subspace
projection. This also makes them more suitable for diagonal covariance
modelling. The HLDA transformation matrix is estimated in maximum
likelihood (ML) framework after building the acoustic models using
39 dimensional MFCCs as described in \cite{hldanagendra}, and is
applied to both feature vectors and the models in the conventional
method.

\section{Experiments and Results}

\subsection{Experimental Setup\label{sub:NMF-Experimental-Setup}}

Aurora-2 task \cite{aurora2} has been used to perform a comparative
study of the proposed techniques versus the existing ones. Aurora-2
consists of connected spoken digit utterances of TIDigits database,
filtered and resampled to 8 kHz, and with noises added at different
SNRs. The noises are sounds recorded in places such as train station,
crowd of people, restaurant, interior of car etc. The availability
of both clean and noisy versions of the training speech utterances
makes them stereo in nature. The test set consists of 10 sets of utterances,
each with one noise environment added, and each at seven distinct
SNR levels.

The acoustic word models for each digit have been built using left
to right continuous density HMMs with 16 states and 3 diagonal covariance
Gaussian mixtures per state. For all the experiments, $C_{0}$ included
MFCC vectors of 13 dimensions, obtained from the signal processing
blocks, are appended with 13 delta and 13 acceleration coefficients
to get a composite 39 dimensional vector per frame. Cepstral mean
subtraction (CMS) has been performed on these vectors, and the resultant
feature vectors are used for building the acoustic models for each
digit, which in Aurora-2 task is a left to right continuous density
HMM with 16 states and 3 diagonal covariance Gaussian mixtures per
state. HMM Toolkit (HTK) 3.4.1 \cite{htkbook} has been used for building
and testing the acoustic models.

NMF has been implemented using MATLAB software. The size of the speech
dictionary (in other words, the dimensionality of the speech subspace)
has been optimised and chosen as $R=20$ throughout the experiments
based on recognition results. It is to be noted that during NMF, LMFB
feature vectors are of dimension $D=23$. While performing NMF, $\mathbf{W}$
has been initialised from random columns of $\mathbf{V}$, and $\mathbf{H}$
with random numbers in $[0,1]$. 500 iterations of NMF are performed
in all the experiments.

For performing HEQ, quantile based method has been employed, dividing
the range of cdf values into 100 quantiles. In the experiments including
HLDA, all 39 directions are retained, since the aim is to observe
only the performance improvement after nullifying the correlations.

\subsection{Results\label{sec:NMF-Results}}

\begin{table}
\protect\caption{Subspace projection on Aurora-2 database\label{tab:NMFresults}}

\begin{centering}
\subfloat[Individual methods\label{tab:NMF-Individual-Methods}]{

\protect\centering{}{\footnotesize{}}%
\begin{tabular}{|c|c|c|c|}
\hline 
\textbf{\footnotesize{}Noise Level} & \textbf{\footnotesize{}Baseline} & \textbf{\footnotesize{}NMF\_plain} & \textbf{\footnotesize{}NMF\_robustW}\tabularnewline
\hline 
\hline 
{\footnotesize{}Clean} & {\footnotesize{}99.25} & {\footnotesize{}99.27} & {\footnotesize{}99.02}\tabularnewline
\hline 
{\footnotesize{}SNR 20} & {\footnotesize{}97.35} & {\footnotesize{}97.50} & {\footnotesize{}97.45}\tabularnewline
\hline 
{\footnotesize{}SNR 15} & {\footnotesize{}93.43} & {\footnotesize{}94.17} & {\footnotesize{}95.00}\tabularnewline
\hline 
{\footnotesize{}SNR 10} & {\footnotesize{}80.62} & {\footnotesize{}82.73} & {\footnotesize{}86.94}\tabularnewline
\hline 
{\footnotesize{}SNR 5} & {\footnotesize{}51.87} & {\footnotesize{}56.48} & {\footnotesize{}65.48}\tabularnewline
\hline 
{\footnotesize{}SNR 0} & {\footnotesize{}24.30} & {\footnotesize{}25.97} & {\footnotesize{}32.46}\tabularnewline
\hline 
{\footnotesize{}SNR -5} & {\footnotesize{}12.03} & {\footnotesize{}12.46} & {\footnotesize{}12.68}\tabularnewline
\hline 
\emph{\footnotesize{}Average} & \emph{\footnotesize{}69.51} & \emph{\footnotesize{}71.37} & \emph{\footnotesize{}75.47}\tabularnewline
\hline 
\end{tabular}\protect{\footnotesize \par}}
\par\end{centering}

\begin{centering}
\subfloat[Cascading with HLDA\label{tab:NMF-Cascading-with-HLDA}]{

\protect\centering{}{\footnotesize{}}%
\begin{tabular}{|c|c|c|c|}
\hline 
\textbf{\footnotesize{}Noise Level} & \textbf{\footnotesize{}HLDA} & \textbf{\footnotesize{}NMF\_plain + HLDA} & \textbf{\footnotesize{}NMF\_robustW + HLDA}\tabularnewline
\hline 
\hline 
{\footnotesize{}Clean} & {\footnotesize{}99.35} & {\footnotesize{}99.35} & {\footnotesize{}99.28}\tabularnewline
\hline 
{\footnotesize{}SNR 20} & {\footnotesize{}98.11} & {\footnotesize{}98.25} & {\footnotesize{}98.26}\tabularnewline
\hline 
{\footnotesize{}SNR 15} & {\footnotesize{}94.84} & {\footnotesize{}95.69} & {\footnotesize{}96.20}\tabularnewline
\hline 
{\footnotesize{}SNR 10} & {\footnotesize{}82.08} & {\footnotesize{}87.14} & {\footnotesize{}89.76}\tabularnewline
\hline 
{\footnotesize{}SNR 5} & {\footnotesize{}49.85} & {\footnotesize{}63.63} & {\footnotesize{}70.95}\tabularnewline
\hline 
{\footnotesize{}SNR 0} & {\footnotesize{}21.64} & {\footnotesize{}28.31} & {\footnotesize{}35.41}\tabularnewline
\hline 
{\footnotesize{}SNR -5} & {\footnotesize{}11.15} & {\footnotesize{}11.47} & {\footnotesize{}12.97}\tabularnewline
\hline 
\emph{\footnotesize{}Average} & \emph{\footnotesize{}69.30} & \emph{\footnotesize{}74.60} & \emph{\footnotesize{}78.11}\tabularnewline
\hline 
\end{tabular}\protect{\footnotesize \par}}
\par\end{centering}

\centering{}\subfloat[Cascading with HEQ\label{tab:NMF-Cascading-with-HEQ}]{

\protect\centering{}{\footnotesize{}}%
\begin{tabular}{|c|c|c|c|}
\hline 
\textbf{\footnotesize{}Noise Level} & \textbf{\footnotesize{}HEQ} & \textbf{\footnotesize{}NMF\_plain + HEQ} & \textbf{\footnotesize{}NMF\_robustW + HEQ}\tabularnewline
\hline 
\hline 
{\footnotesize{}Clean} & {\footnotesize{}99.03} & {\footnotesize{}98.94} & {\footnotesize{}98.73}\tabularnewline
\hline 
{\footnotesize{}SNR 20} & {\footnotesize{}97.68} & {\footnotesize{}97.76} & {\footnotesize{}97.21}\tabularnewline
\hline 
{\footnotesize{}SNR 15} & {\footnotesize{}95.39} & {\footnotesize{}95.93} & {\footnotesize{}95.26}\tabularnewline
\hline 
{\footnotesize{}SNR 10} & {\footnotesize{}90.00} & {\footnotesize{}91.49} & {\footnotesize{}90.67}\tabularnewline
\hline 
{\footnotesize{}SNR 5} & {\footnotesize{}75.61} & {\footnotesize{}79.21} & {\footnotesize{}78.79}\tabularnewline
\hline 
{\footnotesize{}SNR 0} & {\footnotesize{}46.66} & {\footnotesize{}51.63} & {\footnotesize{}53.04}\tabularnewline
\hline 
{\footnotesize{}SNR -5} & {\footnotesize{}18.37} & {\footnotesize{}19.54} & {\footnotesize{}21.28}\tabularnewline
\hline 
\emph{\footnotesize{}Average} & \emph{\footnotesize{}81.07} & \emph{\footnotesize{}83.20} & \emph{\footnotesize{}82.99}\tabularnewline
\hline 
\end{tabular}\protect{\footnotesize \par}}
\end{table}
\begin{figure}
\begin{centering}
\includegraphics[bb=0bp 0bp 439bp 255bp,scale=0.6]{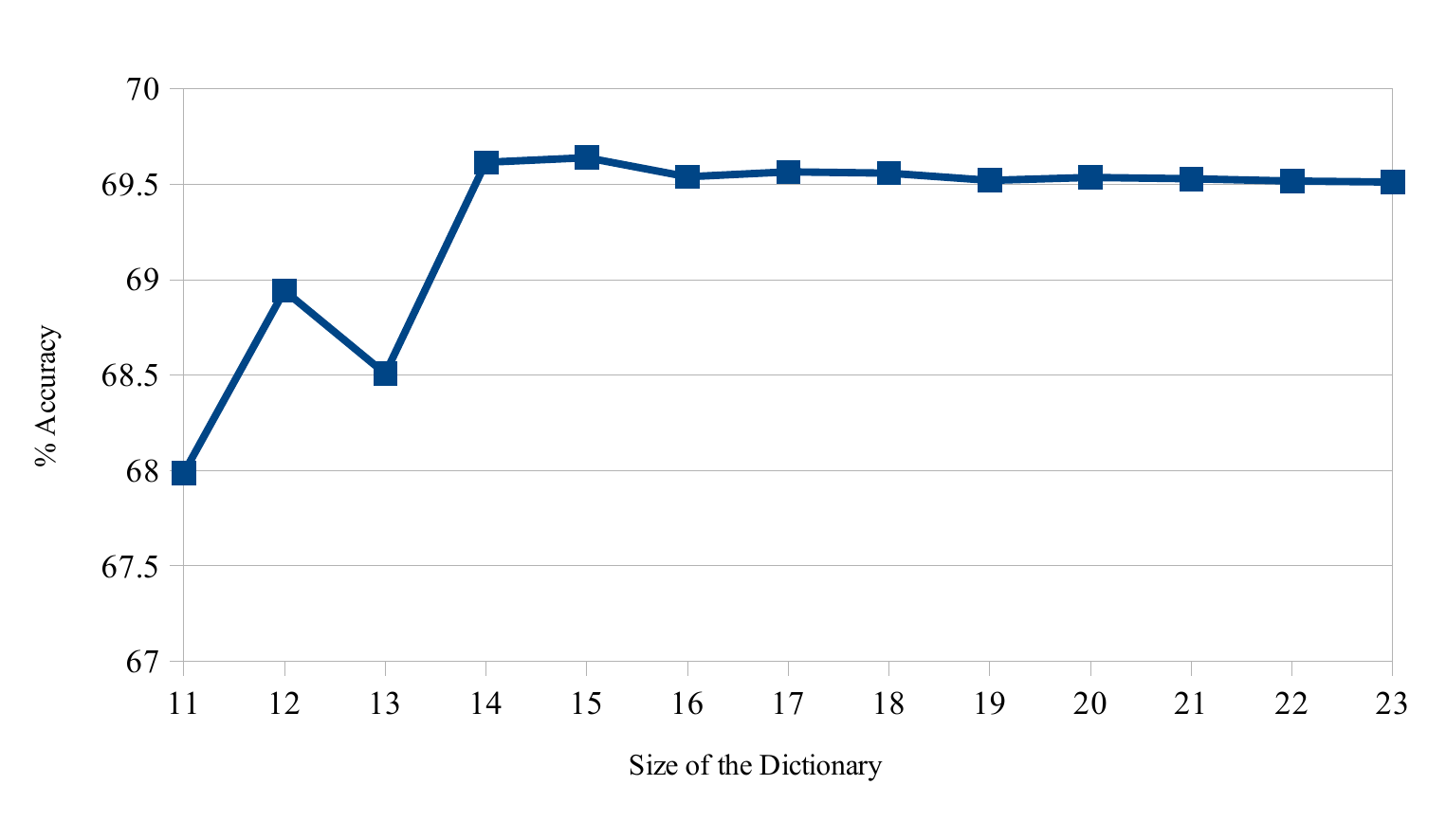}
\par\end{centering}

\protect\caption{Results of PCA as a subspace projection step\label{fig:PCA-Results}}
\end{figure}

Table \ref{tab:NMFresults} shows the recognition accuracies of the
various techniques on Aurora-2 database. Average values shown are
taken over SNR levels $20-0$ dB. Table \ref{tab:NMF-Individual-Methods}
shows the accuracies of proposed methods. Tables \ref{tab:NMF-Cascading-with-HLDA}
and \ref{tab:NMF-Cascading-with-HEQ} show how the methods improve
when cascaded with HLDA and HEQ respectively. It can be observed that
HLDA can be used to achieve noise robustness in low noise levels only,
and fails at high noise levels (SNR $5$ to $-5$ dB). In the PCA
method as shown in Figure \ref{fig:PCA-Results}, a dictionary of
size $23$ is an orthogonal matrix which retains all the directions,
thus corresponding exactly to the baseline system. It can be seen
that this method hardly gives an improvement over the baseline at
different dictionary sizes. NMF\_plain gives $1.86\%$ absolute improvement
in recognition accuracy, NMF\_plain cascaded with HLDA gives $5.09\%$,
NMF\_robustW gives $5.96\%$, NMF\_robustW+HLDA gives $8.6\%$. NMF\_plain+HEQ
gives $13.69\%$ and NMF\_robustW gives $13.48\%$ improvement.

Specifically at SNR 5 dB, the individual proposed methods NMF\_plain
and NMF\_robustW give absolute improvements by $4.61\%$ and $11.76\%$
respectively. When combined with HEQ, the highest improvement achieved
is $27.3\%$ at SNR $5$ and $0$ dB. Figure \ref{fig:PCA-Results}
shows the accuracy of using PCA as a feature processing step at different
sizes of the dictionary.

The combination of NMF\_plain+HEQ+HLDA achieved a recognition accuracy
of $81.26\%$. This is not a comparable improvement over the other
proposed techniques, considering the increased computational complexity.

\section{Discussion\label{sec:Discussion}}

Methods described in Sections \ref{sub:NMF_plain} and \ref{sub:NMF_robustW}
are observed to give improvements in all noise conditions at all SNR
levels, and give significantly better performances in moderate to
high noise conditions. NMF\_robustW+HEQ does not give improvement
over NMF\_plain+HEQ, both of them perform almost equally well. So
when almost equalised weights $\mathbf{H}_{\varphi}$ are obtained
through NMF\_robustW, there is no advantage when an additional equalisation
is done in cepstral domain.

The proposed methods have notable advantages of over other techniques
in literature. There is no need of building a separate dictionary
for capturing noise characteristics, for which training audio files
containing pure noise would have been required. The methods operate
on each speech frame independently, and so they can even handle very
short utterances, unlike HEQ where the estimate of $F_{test}(x)$
will be poor for short utterances. The methods are seen to give advantage
when combined with other feature normalisation techniques. Finally
the proposed methods are simple, easy to implement, and are achievable
without the use of some commonly used additional tools like speech/silence
detector, pre-built CDHMM model etc.

The proposed methods have a few disadvantages. The decomposition is
iterative and the step size of the update rules (\ref{eq:Wupdate})
and (\ref{eq:Hupdate}) is small. So it takes many iterations to converge,
and the number of iterations increases when the size of the database
is large. Dictionary estimation from a very large database is computationally
expensive and is limited by the availability of memory in the computing
device. Iterations have to be performed even during testing, to determine
the weights $\mathbf{H_{\varphi}}$.

Thus the concept of building-blocks representation of speech is incorporated
into the features, still preserving the advantages of using MFCCs.
Conventional HMM based recogniser has been used to test the efficacy
of the proposed MFCCs against the standard MFCCs.

\chapter{Stereo and Non-Stereo Based Feature Compensation\label{chap:SPLICE}}

In this chapter, the techniques based on SPLICE are studied. A simple
modification to SPLICE is proposed, based on an assumption made on
the correlation of training stereo data. This is shown to give improved
performance in all the noise conditions, predominantly in unseen conditions
which are highly mismatched with those of training. This method \emph{does
not need any} \emph{adaptation data}, in contrast to the recent work
proposed in literature \cite{unseensplice}, and has been termed as
\emph{modified SPLICE} (M-SPLICE). M-SPLICE has also been extended
to work for datasets that are not stereo recorded, with minimal performance
degradation as compared to conventional SPLICE. Finally, an MLLR-based
run-time noise adaptation framework has been proposed, which is computationally
efficient and achieves better results than MLLR HMM-adaptation. This
method is done on 13 dimensional MFCCs and does not require two-pass
Viterbi decoding, in contrast to conventional MLLR done on 39 dimensions.

\section{Review of SPLICE}

\label{sec:Review-of-SPLICE}

As discussed in the introduction, SPLICE algorithm makes the following
two assumptions:
\begin{enumerate}
\item The noisy features $\left\{ \mathbf{y}\right\} $ follow a Gaussian
mixture density of $M$ modes
\begin{equation}
p(\mathbf{y})=\sum_{m=1}^{M}P(m)p(\mathbf{y}\,|\,m)=\sum_{m=1}^{M}\boldsymbol{\pi}_{m}\mathcal{N}\left(\mathbf{y}\,;\,\boldsymbol{\mu}_{y,m},\boldsymbol{\Sigma}_{y,m}\right)\label{eq:ubm}
\end{equation}

\item The conditional density $p(\mathbf{x}\,|\,\mathbf{y},m)$ is the Gaussian
\begin{equation}
p(\mathbf{x}\,|\,\mathbf{y},m)\sim\mathcal{N}\left(\mathbf{x}\,;\,\mathbf{A}_{m}\mathbf{y}+\mathbf{b}_{m},\boldsymbol{\Sigma}_{x,m}\right)\label{eq:assum2}
\end{equation}
where $\{\mathbf{x}\}$ are the clean features.
\end{enumerate}
Thus, $\mathbf{A}_{m}$ and $\mathbf{b}_{m}$ parameterise the mixture
specific linear transformations on the noisy vector $\mathbf{y}$.
Here $\mathbf{y}$ and $m$ are independent variables, and $\mathbf{x}$
is dependent on them. Estimate of the \emph{cleaned} feature $\widehat{\mathbf{x}}$
can be obtained in MMSE framework as shown in Eq. (\ref{eq:splice}).

The derivation of SPLICE transformations is briefly discussed next.
Let $\mathbf{W}_{m}=\begin{bmatrix}\mathbf{b}_{m} & \mathbf{A}_{m}\end{bmatrix}$
and $\mathbf{y}'=\begin{bmatrix}1 & \mathbf{y}^{T}\end{bmatrix}^{T}$.
Using $N$ independent pairs of stereo training features $\left\{ \left(\mathbf{x}_{n},\mathbf{y}_{n}\right)\right\} $
and maximising the joint log-likelihood 
\begin{equation}
\mathcal{L}=\sum_{n=1}^{N}\log p(\mathbf{x}_{n},\mathbf{y}_{n})=\sum_{n=1}^{N}\log\left[\sum_{m=1}^{M}p(\mathbf{x}_{n}\,|\,\mathbf{y}_{n},m)p(\mathbf{y}_{n}\,|\,m)P(m)\right]\label{eq:likeli}
\end{equation}
yields
\begin{equation}
\mathbf{W}_{m}=\left[\sum_{n=1}^{N}p\left(m\,|\,\mathbf{y}_{n}\right)\mathbf{x}_{n}\mathbf{y}_{n}'^{T}\right]\left[\sum_{n=1}^{N}p\left(m\,|\,\mathbf{y}_{n}\right)\mathbf{y}_{n}'\mathbf{y}_{n}'^{T}\right]^{-1}\label{eq:splicew}
\end{equation}

Alternatively, sub-optimal update rules of separately estimating $\mathbf{b}_{m}$
and $\mathbf{A}_{m}$ can be derived by initially assuming $\mathbf{A}_{m}$
to be identity matrix while estimating $\mathbf{b}_{m}$. The newly
estimated $\mathbf{b}_{m}$ is then used to estimate $\mathbf{A}_{m}$.

A perfect correlation between $\mathbf{x}$ and $\mathbf{y}$ is assumed,
and the following approximation is used in deriving Eq. (\ref{eq:splicew})
\cite{ssmafify}. 
\begin{equation}
p\left(m\,|\,\mathbf{x}_{n},\mathbf{y}_{n}\right)\approx p\left(m\,|\,\mathbf{x}_{n}\right)\approx p\left(m\,|\,\mathbf{y}_{n}\right)\label{eq:perfcorr}
\end{equation}

Given mixture index $m$, Eq. (\ref{eq:splicew}) can be shown to
give the MMSE estimator of $\widehat{\mathbf{x}}_{m}=\mathbf{A}_{m}\mathbf{y}+\mathbf{b}_{m}$
\cite{lideng}, given by 
\begin{equation}
\widehat{\mathbf{x}}_{m}=\boldsymbol{\mu}_{x,m}+\boldsymbol{\Sigma}_{xy,m}\boldsymbol{\Sigma}_{y,m}^{-1}\left(\mathbf{y}-\boldsymbol{\mu}_{y,m}\right)\label{eq:mmsee}
\end{equation}
where 
\begin{equation}
\boldsymbol{\mu}_{x,m}=\frac{\underset{n=1}{\overset{N}{\sum}}p\left(m\,|\,\mathbf{y}_{n}\right)\mathbf{x}_{n}}{\underset{n=1}{\overset{N}{\sum}}p\left(m\,|\,\mathbf{y}_{n}\right)},\;\boldsymbol{\mu}_{y,m}=\frac{\underset{n=1}{\overset{N}{\sum}}p\left(m\,|\,\mathbf{y}_{n}\right)\mathbf{y}_{n}}{\underset{n=1}{\overset{N}{\sum}}p\left(m\,|\,\mathbf{y}_{n}\right)}\label{eq:SPLICEmeans}
\end{equation}
\begin{equation}
\boldsymbol{\Sigma}_{xy,m}=\frac{\underset{n=1}{\overset{N}{\sum}}p\left(m\,|\,\mathbf{y}_{n}\right)\mathbf{x}_{n}\mathbf{y}_{n}^{T}}{\underset{n=1}{\overset{N}{\sum}}p\left(m\,|\,\mathbf{y}_{n}\right)},\;\boldsymbol{\Sigma}_{y,m}=\frac{\underset{n=1}{\overset{N}{\sum}}p\left(m\,|\,\mathbf{y}_{n}\right)\mathbf{y}_{n}\mathbf{y}_{n}^{T}}{\underset{n=1}{\overset{N}{\sum}}p\left(m\,|\,\mathbf{y}_{n}\right)}\label{eq:SPLICEcov}
\end{equation}
i.e., the alignments $p(m\,|\,\mathbf{y}_{n})$ are being used in
place of $p(m\,|\,\mathbf{x}_{n})$ and $p(m\,|\,\mathbf{x}_{n},\mathbf{y}_{n})$
in Eqs. (\ref{eq:SPLICEmeans}) and (\ref{eq:SPLICEcov}) respectively.
Thus from (\ref{eq:mmsee}),
\begin{equation}
\mathbf{A}_{m}=\boldsymbol{\Sigma}_{xy,m}\boldsymbol{\Sigma}_{y,m}^{-1}\label{eq:ammse}
\end{equation}
\begin{equation}
\mathbf{b}_{m}=\boldsymbol{\mu}_{x,m}-\mathbf{A}_{m}\boldsymbol{\mu}_{y,m}\label{eq:bmmse}
\end{equation}

To reduce the number of parameters, a simplified model with only bias
$\mathbf{b}_{m}$ is proposed in literature \cite{lideng}.

A diagonal version of Eq. (\ref{eq:mmsee}) can be written as
\begin{equation}
\widehat{x}_{c}=\mu_{x,c}+\frac{\sigma_{xy,c}^{2}}{\sigma_{y,c}^{2}}\left(y-\mu_{y,c}\right)\label{eq:diagsplice}
\end{equation}
where $c$ runs along all components of the features and all mixtures.
Since this method does not capture all the correlations, it suffers
from performance degradation. This shows that noise has significant
effect on feature correlations.

\section{Proposed Modification to SPLICE}

\label{sec:Proposed-Modification}

SPLICE assumes that a perfect correlation exists between clean and
noisy stereo features (Eq. (\ref{eq:perfcorr})), which makes the
implementation simple \cite{ssmafify}. But, the actual feature correlations
$\boldsymbol{\Sigma}_{xy,m}$ are used to train SPLICE parameters,
as seen in Eq. (\ref{eq:ammse}). Instead, if the training process
also assumes perfect correlation and eliminates the term $\boldsymbol{\Sigma}_{xy,m}$
during parameter estimation, it complies with the assumptions and
gives improved performance. This simple modification can be done as
follows:

Eq. (\ref{eq:diagsplice}) can be rewritten as
\[
\frac{\widehat{x}-\mu_{x}}{\sigma_{x}}=\frac{\sigma_{xy}^{2}}{\sigma_{x}\sigma_{y}}\left(\frac{y-\mu_{y}}{\sigma_{y}}\right)=\rho\left(\frac{y-\mu_{y}}{\sigma_{y}}\right)
\]
where $\rho=\frac{\sigma_{xy}^{2}}{\sigma_{x}\sigma_{y}}$ is the
correlation coefficient. A perfect correlation implies $\rho=1$.
Since Eq. (\ref{eq:perfcorr}) makes this assumption, it can be enforced
in the above equation to obtain
\begin{equation}
\widehat{x}_{c}=\mu_{x,c}+\frac{\sigma_{x,c}}{\sigma_{y,c}}\left(y-\mu_{y,c}\right)\label{eq:diagcmsplice}
\end{equation}

Similarly, for multidimensional case, the matrix $\boldsymbol{\Sigma}_{x,m}^{-\frac{1}{2}}\boldsymbol{\Sigma}_{xy,m}\boldsymbol{\Sigma}_{y,m}^{-\frac{1}{2}}$
should be enforced to be identity as per the assumption. Thus, the
following is obtained:
\begin{equation}
\widehat{\mathbf{x}}_{m}=\boldsymbol{\mu}_{x,m}+\boldsymbol{\Sigma}_{x,m}^{\frac{1}{2}}\boldsymbol{\Sigma}_{y,m}^{-\frac{1}{2}}\left(\mathbf{y}-\boldsymbol{\mu}_{y,m}\right)\label{eq:modsplice}
\end{equation}

Hence M-SPLICE and its updates are defined as 
\begin{equation}
\widehat{\mathbf{x}}=\sum_{m=1}^{M}p\left(m\,|\,\mathbf{y}\right)\left(\mathbf{C}_{m}\mathbf{y}+\mathbf{d}_{m}\right)\label{eq:modsplicex}
\end{equation}
\begin{align}
\mathbf{C}_{m} & =\boldsymbol{\Sigma}_{x,m}^{\frac{1}{2}}\boldsymbol{\Sigma}_{y,m}^{-\frac{1}{2}}\label{eq:cm}\\
\mathbf{d}_{m} & =\boldsymbol{\mu}_{x,m}-\mathbf{C}_{m}\boldsymbol{\mu}_{y,m}\label{eq:dm}
\end{align}

All the assumptions of conventional SPLICE are valid for M-SPLICE.
Comparing both the methods, it can be seen from Eqs. (\ref{eq:mmsee})
and (\ref{eq:cm}) that while $\mathbf{A}_{m}$ is obtained using
MMSE estimation framework, $\mathbf{C}_{m}$ is based on whitening
expression. Also, $\mathbf{A}_{m}$ involves cross-covariance term
$\boldsymbol{\Sigma}_{xy,m}$, whereas $\mathbf{C}_{m}$ does not.
The bias terms are computed in the same manner, using their respective
transformation matrices, as seen in Eqs. (\ref{eq:bmmse}) and (\ref{eq:dm}).
More analysis on M-SPLICE is given in Section \ref{sub:Motivation-Non-Stereo}.
\begin{figure}
\begin{centering}
\subfloat[M-SPLICE\label{fig:BD-M-SPLICE}]{\protect\begin{centering}
\protect\includegraphics[scale=0.4]{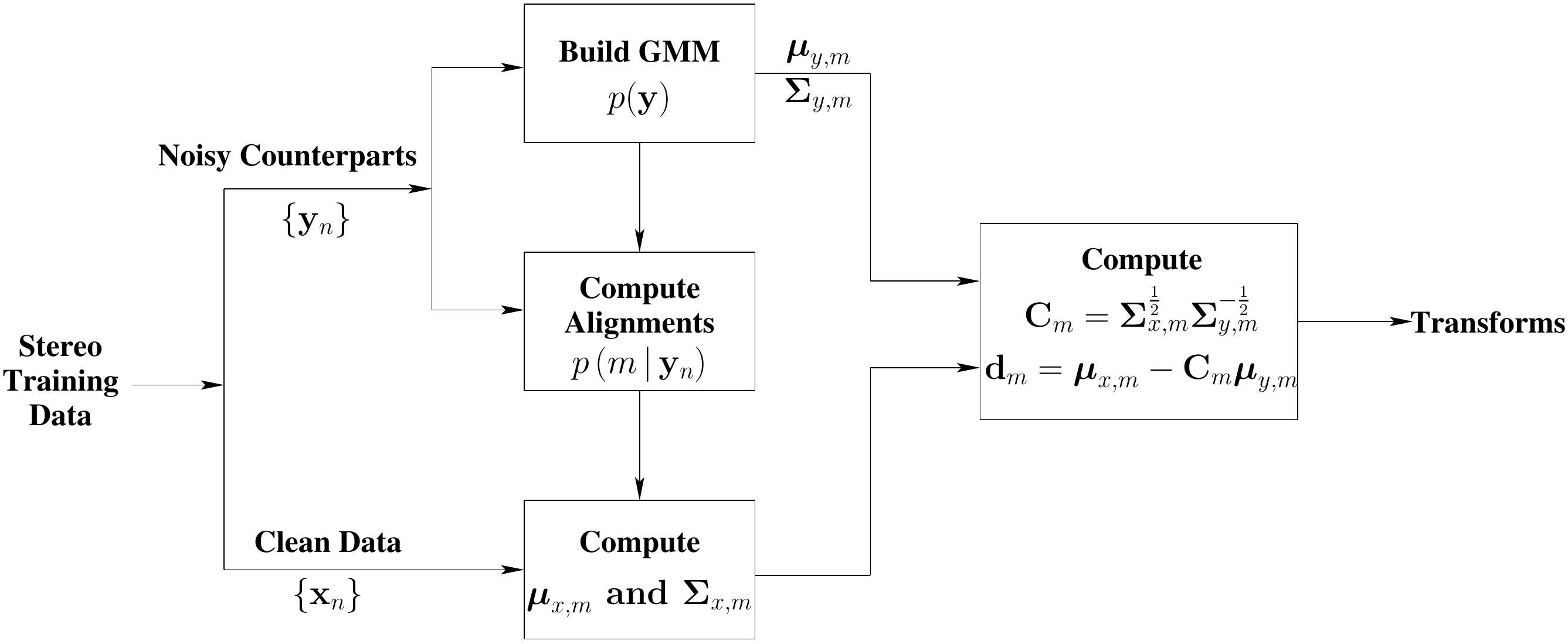}\protect
\par\end{centering}

}
\par\end{centering}

\centering{}\subfloat[Non-Stereo Method\label{fig:BD-Non-Stereo-Method}]{\protect\begin{centering}
\protect\includegraphics[scale=0.4]{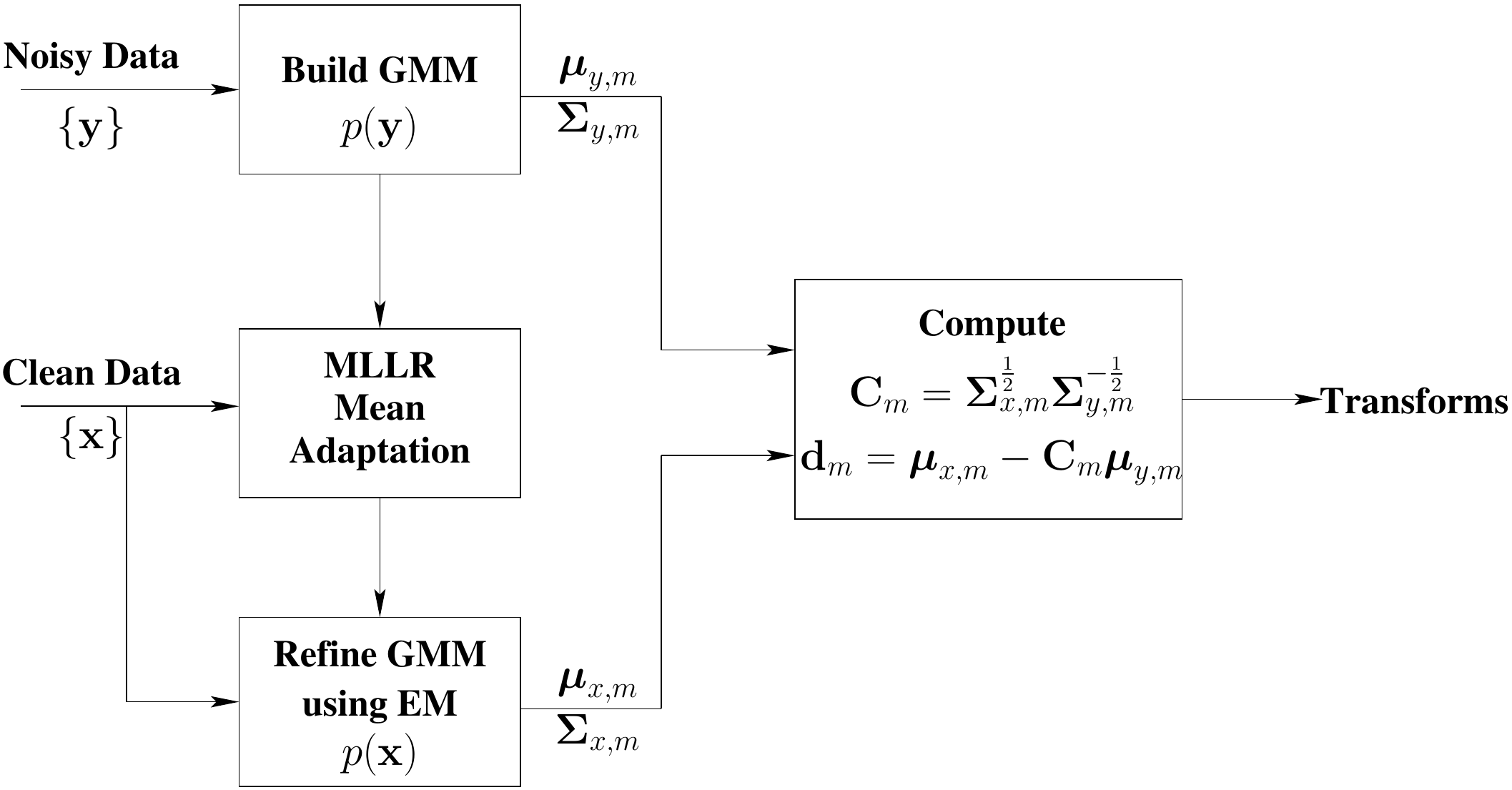}\protect
\par\end{centering}

}\protect\caption{Transform estimation block diagrams of proposed methods\label{fig:BD-Estimation-transforms}}
\end{figure}

\subsection{Training}

The estimation procedure of M-SPLICE transformations is shown in Figure
\ref{fig:BD-M-SPLICE}. The steps are summarised as follows:
\begin{enumerate}
\item Build noisy GMM\footnote{A non-standard term \emph{noisy mixture} has been used to denote a
Gaussian mixture built using noisy data. Similar meanings apply to
\emph{clean mixture},\emph{ noisy GMM} and \emph{clean GMM}.} $p(\mathbf{y})$ using noisy features $\{\mathbf{y}_{n}\}$ of stereo
data. This gives $\boldsymbol{\mu}_{y,m}$ and $\boldsymbol{\Sigma}_{y,m}$.
\item For every noise frame $\mathbf{y}_{n}$, compute the alignment w.r.t.
the noisy GMM, i.e., $p(m\,|\,\mathbf{y}_{n})$.
\item Using the alignments of stereo counterparts, compute the means $\boldsymbol{\mu}_{x,m}$
and covariance matrices $\boldsymbol{\Sigma}_{x,m}$ of each clean
mixture from clean data $\{\mathbf{x}_{n}\}$.
\item Compute $\mathbf{C}_{m}$ and $\mathbf{d}_{m}$ using Eq. (\ref{eq:cm})
and (\ref{eq:dm}).
\end{enumerate}

\subsection{Testing\label{sub:Testing-M-SPLICE}}

Testing process of M-SPLICE is exactly same as that of conventional
SPLICE, and is summarised as follows:
\begin{enumerate}
\item For each test vector $\mathbf{y}$, compute the alignment w.r.t. the
noisy GMM, i.e., $p(m\,|\,\mathbf{y})$.
\item Compute the cleaned version as:
\[
\widehat{\mathbf{x}}=\sum_{m=1}^{M}p\left(m\,|\,\mathbf{y}\right)\left(\mathbf{C}_{m}\mathbf{y}+\mathbf{d}_{m}\right)
\]

\end{enumerate}

\subsection{M-SPLICE with Diagonal Transformations}

Techniques such as CMS, HEQ etc. operate on individual feature dimensions,
assuming the features have diagonal covariance structures. This assumption
is valid for MFCCs, since the use of DCT approximately decorrelates
the features. Without significant loss of performance, M-SPLICE can
also be extended in a similar fashion by constraining the covariance
matrices $\boldsymbol{\Sigma}_{x,m}$ and $\boldsymbol{\Sigma}_{y,m}$
to be diagonal. Thus $\mathbf{C}_{m}$ becomes diagonal, and Eq. \ref{eq:modsplicex}
can be rewritten as
\[
\widehat{\mathbf{x}}=\sum_{m=1}^{M}p\left(m\,|\,\mathbf{y}\right)\left(\mathbf{c}_{m}^{T}\mathbf{y}+\mathbf{d}_{m}\right)
\]
where $diag\left(\mathbf{c}_{m}\right)=\mathbf{C}_{m}$. This implementation
replaces the matrix multiplication in M-SPLICE by scalar product and
addition operations.

\section{Non-Stereo Extension}

\label{sec:Non-Stereo-Extension}This section motivates and proposes
the extension of M-SPLICE to datasets which are not stereo recorded.
However some noisy training utterances, which are not necessarily
the stereo counterparts of the clean data, are required.

\subsection{Motivation\label{sub:Motivation-Non-Stereo}}

Consider a stereo dataset of $N$ training frames $(\mathbf{x}_{n},\mathbf{y}_{n})$.
Suppose two $M$ mixture GMMs $p(\mathbf{x})$ and $p(\mathbf{y})$
are independently built using $\{\mathbf{x}_{n}\}$ and $\{\mathbf{y}_{n}\}$
respectively, and each data point is hard-clustered to the mixture
giving the highest probability. The matrix $\mathbf{V}_{M\times M}$,
built as described below, is of interest: 
\[
\mathbf{V}_{ij}=\sum_{n=1}^{N}\mathbbm{1}\left(\mathbf{x}_{n}\in i,\mathbf{y}_{n}\in j\right)
\]
where $\mathbbm{1()}$ is indicator function. In other words, while
parsing the stereo training data, when a stereo pair with clean part
belonging to $i^{th}$ clean mixture and noisy part to $j^{th}$ noisy
mixture is encountered, the $ij^{th}$ element of the matrix is incremented
by unity. Thus each $ij^{th}$ element of the matrix denotes the number
of stereo pairs belong to the $i^{th}$ clean $-$ $j^{th}$ noisy
mixture-pair. When data are soft assigned to all the mixtures, the
matrix can instead be built as:
\[
\mathbf{V}_{ij}=\sum_{n=1}^{N}p(i\,|\,\mathbf{x}_{n})p(j\,|\,\mathbf{y}_{n})
\]

\begin{figure*}
\noindent \begin{centering}
\subfloat[Separately built noisy and clean GMMs\label{fig:Sky}]{\protect\begin{centering}
\protect\includegraphics[scale=0.36]{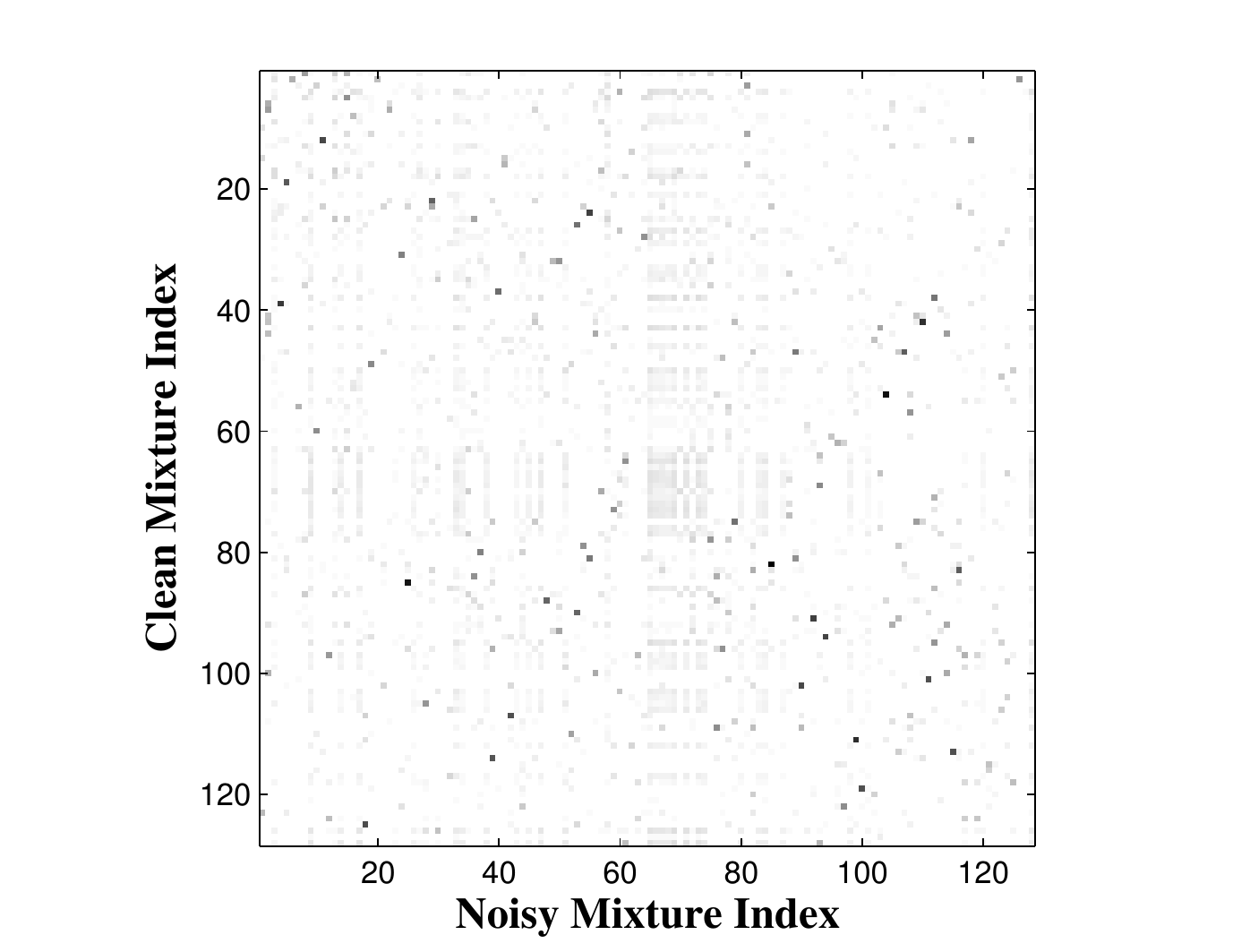}\protect
\par\end{centering}

}\subfloat[GMMs of SPLICE and M-SPLICE\label{fig:SPLICE-M-SPLICE-Sky}]{\protect\begin{centering}
\protect\includegraphics[scale=0.36]{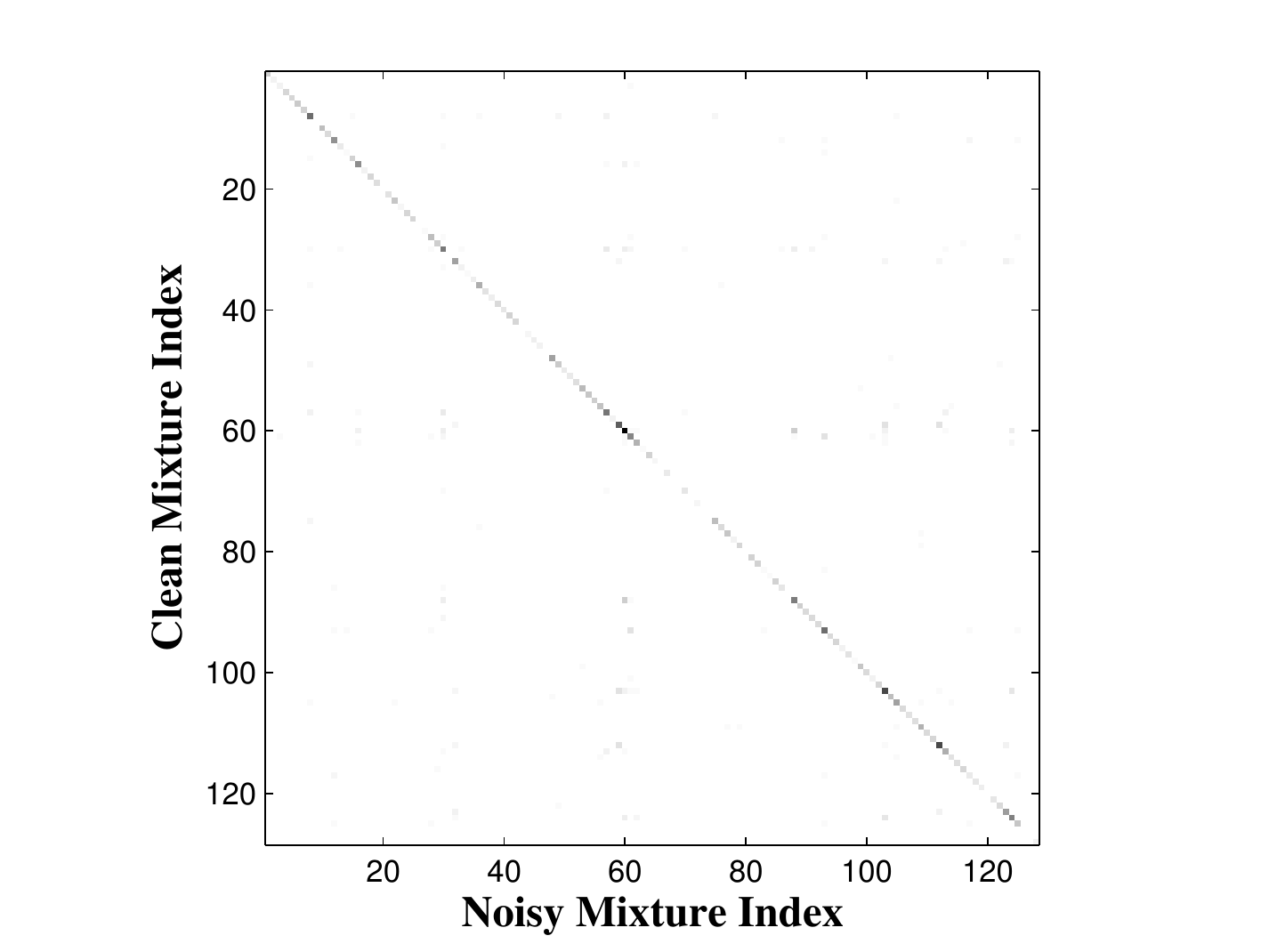}\protect
\par\end{centering}

}\subfloat[Noisy GMM and MLLR-EM based clean GMM\label{fig:Non-Stereo-Sky}]{\protect\begin{centering}
\protect\includegraphics[scale=0.36]{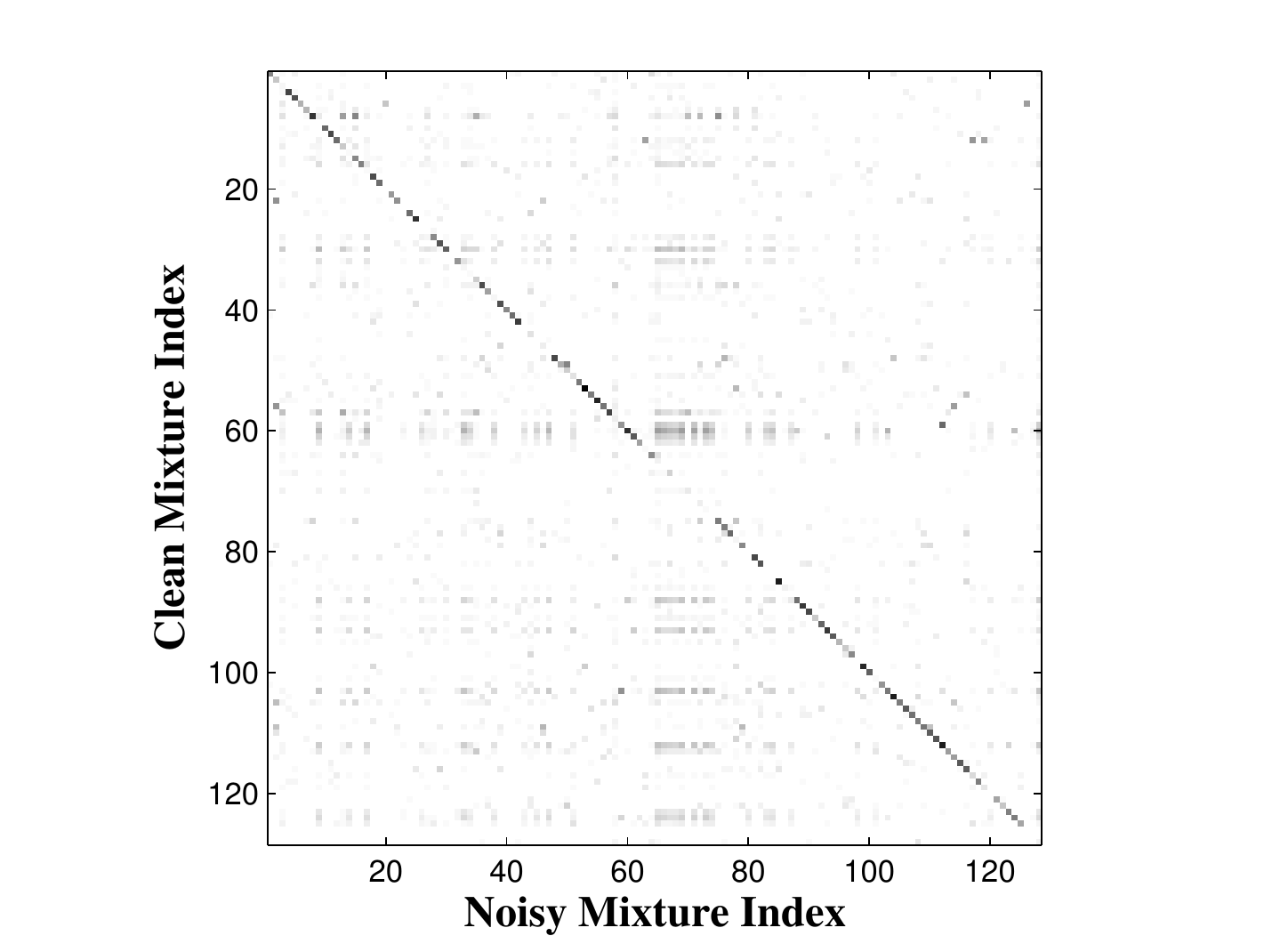}\protect
\par\end{centering}

}
\par\end{centering}

\protect\caption{Mixture assignment distribution plots for Aurora-2 stereo training
data\label{fig:Class-label-distribution-plots}}
\end{figure*}
Figure \ref{fig:Sky} visualises such a matrix built using Aurora-2
stereo training data using $128$ mixture models. A dark spot in the
plot represents a higher data count, and a bulk of stereo data points
do belong to that mixture-pair. 

In conventional SPLICE and M-SPLICE, only the noisy GMM $p(\mathbf{y})$
is built, and not $p(\mathbf{x})$. $p\left(m\,|\,\mathbf{y}_{n}\right)$
are computed for every noisy frame, and the same alignments are assumed
for the clean frames $\{\mathbf{x}_{n}\}$ while computing $\boldsymbol{\mu}_{x,m}$
and \foreignlanguage{english}{$\boldsymbol{\Sigma}_{x,m}$}. Hence
$\boldsymbol{\mu}_{x,m}$, \foreignlanguage{english}{$\boldsymbol{\Sigma}_{x,m}$}
and $p\left(m\,|\,\mathbf{y}\right)$ can be considered as the parameters
of a clean hypothetical GMM $p(\mathbf{x})$. Now, given these GMMs
$p(\mathbf{y})$ and $p(\mathbf{x})$, the matrix $\mathbf{V}$ can
be constructed, which is visualised in Fig. (\ref{fig:SPLICE-M-SPLICE-Sky}).
Since the alignments are same, and $i^{th}$ clean mixture corresponds
to the $i^{th}$ noisy mixture, a diagonal pattern can be seen.

Thus, under the assumption of Eq. (\ref{eq:perfcorr}), conventional
SPLICE and M-SPLICE are able to estimate transforms from $i^{th}$
noisy mixture to exactly $i^{th}$ clean mixture by maintaining the
mixture-correspondence.

When stereo  not available, such exact mixture correspondence do not
exist. Fig. \ref{fig:Sky} makes this fact evident, since stereo property
was not used while building the two independent GMMs. However, a sparse
structure can be seen, which suggests that for most noisy mixtures
$j$, there exists a unique clean mixture $i^{*}$ having highest
mixture-correspondence. This property can be exploited to estimate
piecewise linear transformations from every mixture $j$ of $p(\mathbf{y})$
to a single mixture $i^{*}$ of $p(\mathbf{x})$, ignoring all other
mixtures $i\neq i^{*}$. This is the basis for the proposed extension
to non-stereo data.

\subsection{Implementation}

In the absence of stereo data, the approach is to build two separate
GMMs viz., clean and noisy during training, such that there exists
mixture-to-mixture correspondence between them, as close to Fig. \ref{fig:SPLICE-M-SPLICE-Sky}
as possible. Then whitening based transforms can be estimated from
each noisy mixture to its corresponding clean mixture. This sort of
extension is not obvious in the conventional SPLICE framework, since
it is not straight-forward to compute the cross-covariance terms $\boldsymbol{\Sigma}_{xy,m}$
without using stereo data. Also, M-SPLICE is expected to work better
than SPLICE due to its advantages described earlier.

The training approach of two mixture-corresponded GMMs is as follows:
\begin{enumerate}
\item After building the noisy GMM $p(\mathbf{y})$, it is mean adapted
by estimating a global MLLR transformation using clean training data.
The transformed GMM has the same covariances and weights, and only
means are altered to match the clean data. By this process, the mixture
correspondences are not lost.
\item However, the transformed GMM need not model the clean data accurately.
So a few (typically three) steps of expectation maximisation (EM)
are performed using clean training data, initialising with the transformed
GMM. This adjusts all the parameters and gives a more accurate representation
of the clean GMM $p(\mathbf{x})$.
\end{enumerate}
Now, the matrix obtained through this method using Aurora-2 training
data is visualised in Figure \ref{fig:Non-Stereo-Sky}. It can be
noted that no stereo information has been used while obtaining $p(\mathbf{x})$,
following the above mentioned steps, from $p(\mathbf{y})$. It can
be observed that a diagonal pattern is retained, as in the case of
M-SPLICE, though there are some outliers. Since stereo information
is not used, only comparable performances can be achieved. Figure
\ref{fig:BD-Non-Stereo-Method} shows the block diagram of estimating
transformations of non-stereo method. The steps are summarised as
follows:
\begin{enumerate}
\item Build noisy GMM $p(\mathbf{y})$ using noisy features $\{\mathbf{y}\}$.
This gives $\boldsymbol{\mu}_{y,m}$ and $\boldsymbol{\Sigma}_{y,m}$.
\item Adapt the means of noisy GMM $p(\mathbf{y})$ to clean data $\{\mathbf{x}\}$
using global MLLR transformation.
\item Perform at least three EM iterations to refine the adapted GMM using
clean data. This gives $p(\mathbf{x})$, thus $\boldsymbol{\mu}_{x,m}$
and $\boldsymbol{\Sigma}_{x,m}$.
\item Compute $\mathbf{C}_{m}$ and $\mathbf{d}_{m}$ using Eq. (\ref{eq:cm})
and (\ref{eq:dm}).
\end{enumerate}
The testing process is exactly same as that of M-SPLICE, as explained
in Section \ref{sub:Testing-M-SPLICE}.

\section{Additional Run-time Adaptation}

\label{sec:Additional-Run-time-Adaptation}To improve the performance
of the proposed methods during run-time, GMM adaptation to the test
condition can be done in both conventional SPLICE and M-SPLICE frameworks
in a simple manner. Conventional MLLR adaptation on HMMs involves
two-pass recognition, where the transformation matrices are estimated
using the alignments obtained through first pass Viterbi-decoded output,
and a final recognition is performed using the transformed models.

MLLR adaptation can be used to adapt GMMs in the context of SPLICE
and M-SPLICE as follows:
\begin{enumerate}
\item Adapt the noisy GMM through a global MLLR mean transformation
\[
\boldsymbol{\mu}_{y,m}^{(a)}\leftarrow\boldsymbol{\mu}_{y,m}
\]

\item Now, adjust the bias term in conventional SPLICE or M-SPLICE as
\begin{equation}
\mathbf{d}_{m}^{(a)}=\boldsymbol{\mu}_{x,m}-\mathbf{C}_{m}\boldsymbol{\mu}_{y,m}^{(a)}\label{eq:run-time-adapt}
\end{equation}

\end{enumerate}
This method involves only simple calculation of alignments of the
test data w.r.t. the noisy GMM, and doesn't need Viterbi decoding.
Clean mixture means $\mathbf{\mu}_{x,m}$ computed during training
need to be stored. A separate global MLLR mean transform can be estimated
using test utterances belonging to each noise condition. The steps
for testing process for run-time compensation are summarised as follows:
\begin{enumerate}
\item For all test vectors $\{\mathbf{y}\}$ belonging to a particular environment,
compute the alignments w.r.t. the noisy GMM, i.e., $p(m\,|\,\mathbf{y})$.
\item Estimate a global MLLR mean transformation using $\{\mathbf{y}\}$,
maximising the likelihood w.r.t. $p(\mathbf{y})$.
\item Compute the adapted noisy GMM $p^{(a)}(\mathbf{y})$ using the estimated
MLLR transform. Only the means $\boldsymbol{\mu}_{y,m}$ of the noisy
GMM would have been adapted as $\boldsymbol{\mu}_{y,m}^{(a)}$.
\item Using Eq. (\ref{eq:run-time-adapt}), recompute the bias term of SPLICE
or M-SPLICE.
\item Compute the cleaned test vectors as 
\[
\widehat{\mathbf{x}}=\sum_{m=1}^{M}p\left(m\,|\,\mathbf{y}\right)\left(\mathbf{C}_{m}\mathbf{y}+\mathbf{d}_{m}^{(a)}\right)
\]

\end{enumerate}

\section{Experiments and Results}

\subsection{Experimental Setup}

All SPLICE based linear transformations have been applied on 13 dimensional
MFCCs, including $C_{0}$. Aurora-2 setup is the same as described
in \ref{sub:NMF-Experimental-Setup}. During HMM training, the features
are appended with 13 delta and 13 acceleration coefficients to get
a composite 39 dimensional vector per frame. Cepstral mean subtraction
(CMS) has been performed in all the experiments. 128 mixture GMMs
are built for all SPLICE based experiments. Run-time noise adaptation
in SPLICE framework is performed on 13 dimensional MFCCs. Data belonging
to each SNR level of a test noise condition has been separately used
to compute the global transformations. In all SPLICE based experiments,
pseudo-cleaning of clean features has been performed.

To test the efficacy of non-stereo method on a database which does
not contain stereo data, Aurora-4 task of 8 kHz sampling frequency
has been used. Aurora-4 is a continuous speech recognition task with
clean and noisy training utterances (non-stereo) and test utterances
of 14 environments. Aurora-4 acoustic models are built using crossword
triphone HMMs of 3 states and 6 mixtures per state. Standard WSJ0
bigram language model has been used during decoding of Aurora-4. Noisy
GMM of 512 mixtures is built for evaluating non-stereo method, using
7138 utterances taken from both clean and multi-training data. This
GMM is adapted to standard clean training set to get the clean GMM.

\subsection{Results}

\label{sec:Experimental-Results}
\begin{table}
\noindent \begin{centering}
\protect\caption{SPLICE-based methods on Aurora-2 database\label{tab:SPLICE-Aurora2}}
\subfloat[Individual Methods\label{tab:Individual-Methods}]{\protect\centering{}{\footnotesize{}}%
\begin{tabular}{|c|c|c|c|>{\centering}p{1.8cm}|>{\centering}p{1.8cm}|>{\centering}p{2.8cm}|}
\hline 
{\footnotesize{}Noise Level} & {\footnotesize{}Baseline} & {\footnotesize{}SPLICE} & {\footnotesize{}M-SPLICE} & {\footnotesize{}M-SPLICE }\emph{\footnotesize{}Diagonal} & {\footnotesize{}Non-Stereo Method} & {\footnotesize{}Non-Stereo Method }\emph{\footnotesize{}Diagonal}\tabularnewline
\hline 
\hline 
{\footnotesize{}Clean} & {\footnotesize{}99.25} & {\footnotesize{}98.97} & {\footnotesize{}99.01} & {\footnotesize{}98.98} & {\footnotesize{}99.08} & {\footnotesize{}99.03}\tabularnewline
\hline 
{\footnotesize{}SNR 20} & {\footnotesize{}97.35} & {\footnotesize{}97.84} & {\footnotesize{}97.92} & {\footnotesize{}97.85} & {\footnotesize{}97.68} & {\footnotesize{}97.67}\tabularnewline
\hline 
{\footnotesize{}SNR 15} & {\footnotesize{}93.43} & {\footnotesize{}95.81} & {\footnotesize{}96.10} & {\footnotesize{}95.78} & {\footnotesize{}95.15} & {\footnotesize{}95.01}\tabularnewline
\hline 
{\footnotesize{}SNR 10} & {\footnotesize{}80.62} & {\footnotesize{}89.48} & {\footnotesize{}91.03} & {\footnotesize{}90.19} & {\footnotesize{}87.37} & {\footnotesize{}86.74}\tabularnewline
\hline 
{\footnotesize{}SNR 5} & {\footnotesize{}51.87} & {\footnotesize{}72.71} & {\footnotesize{}77.59} & {\footnotesize{}75.46} & {\footnotesize{}68.49} & {\footnotesize{}67.35}\tabularnewline
\hline 
{\footnotesize{}SNR 0} & {\footnotesize{}24.30} & {\footnotesize{}42.85} & {\footnotesize{}50.72} & {\footnotesize{}46.60} & {\footnotesize{}39.00} & {\footnotesize{}37.76}\tabularnewline
\hline 
{\footnotesize{}SNR -5} & {\footnotesize{}12.03} & {\footnotesize{}18.52} & {\footnotesize{}22.27} & {\footnotesize{}19.39} & {\footnotesize{}16.73} & {\footnotesize{}16.31}\tabularnewline
\hline 
\hline 
{\footnotesize{}Test A} & {\footnotesize{}67.45} & {\footnotesize{}81.39} & {\footnotesize{}83.47} & {\footnotesize{}81.72} & {\footnotesize{}77.44} & {\footnotesize{}76.64}\tabularnewline
\hline 
{\footnotesize{}Test B} & {\footnotesize{}72.26} & {\footnotesize{}83.24} & {\footnotesize{}84.18} & {\footnotesize{}82.43} & {\footnotesize{}79.63} & {\footnotesize{}79.06}\tabularnewline
\hline 
{\footnotesize{}Test C} & {\footnotesize{}68.14} & {\footnotesize{}69.42} & {\footnotesize{}78.06} & {\footnotesize{}77.57} & {\footnotesize{}73.54} & {\footnotesize{}73.13}\tabularnewline
\hline 
\emph{\footnotesize{}Overall} & \emph{\footnotesize{}69.51} & \emph{\footnotesize{}79.74} & \emph{\footnotesize{}82.67} & \emph{\footnotesize{}81.17} & \emph{\footnotesize{}77.54} & \emph{\footnotesize{}76.91}\tabularnewline
\hline 
\end{tabular}\protect{\footnotesize \par}}
\par\end{centering}

\noindent \centering{}\subfloat[Run-time adaptation\label{tab:Run-time-adaptation}]{\protect\begin{centering}
\protect
\par\end{centering}

\protect\centering{}{\footnotesize{}}%
\begin{tabular}{|c|>{\centering}p{1.1cm}|>{\centering}p{1.3cm}|>{\centering}p{1.7cm}|>{\centering}p{1.7cm}|>{\centering}p{1.8cm}|>{\centering}p{2.5cm}|}
\hline 
{\footnotesize{}Noise Level} & {\footnotesize{}MLLR (39)} & {\footnotesize{}SPLICE + RA} & {\footnotesize{}M-SPLICE + RA} & {\footnotesize{}M-SPLICE }\emph{\footnotesize{}Diagonal}{\footnotesize{}
+ RA} & {\footnotesize{}Non-Stereo Method + RA} & {\footnotesize{}Non-Stereo Method }\emph{\footnotesize{}Diagonal}{\footnotesize{}
+ RA}\tabularnewline
\hline 
\hline 
{\footnotesize{}Clean} & {\footnotesize{}99.28} & {\footnotesize{}99.05} & {\footnotesize{}99.02} & {\footnotesize{}99.00} & {\footnotesize{}99.08} & {\footnotesize{}99.07}\tabularnewline
\hline 
{\footnotesize{}SNR 20} & {\footnotesize{}98.33} & {\footnotesize{}97.96} & {\footnotesize{}98.18} & {\footnotesize{}98.22} & {\footnotesize{}97.77} & {\footnotesize{}97.75}\tabularnewline
\hline 
{\footnotesize{}SNR 15} & {\footnotesize{}96.82} & {\footnotesize{}96.21} & {\footnotesize{}96.87} & {\footnotesize{}96.70} & {\footnotesize{}95.47} & {\footnotesize{}95.38}\tabularnewline
\hline 
{\footnotesize{}SNR 10} & {\footnotesize{}91.88} & {\footnotesize{}90.61} & {\footnotesize{}93.10} & {\footnotesize{}92.61} & {\footnotesize{}88.80} & {\footnotesize{}88.77}\tabularnewline
\hline 
{\footnotesize{}SNR 5} & {\footnotesize{}73.88} & {\footnotesize{}75.05} & {\footnotesize{}82.00} & {\footnotesize{}81.05} & {\footnotesize{}72.36} & {\footnotesize{}72.34}\tabularnewline
\hline 
{\footnotesize{}SNR 0} & {\footnotesize{}41.94} & {\footnotesize{}46.27} & {\footnotesize{}57.51} & {\footnotesize{}55.70} & {\footnotesize{}44.98} & {\footnotesize{}44.84}\tabularnewline
\hline 
{\footnotesize{}SNR -5} & {\footnotesize{}18.71} & {\footnotesize{}20.10} & {\footnotesize{}27.32} & {\footnotesize{}26.30} & {\footnotesize{}20.43} & {\footnotesize{}20.27}\tabularnewline
\hline 
\hline 
{\footnotesize{}Test A} & {\footnotesize{}79.31} & {\footnotesize{}82.45} & {\footnotesize{}86.47} & {\footnotesize{}85.90} & {\footnotesize{}80.12} & {\footnotesize{}80.01}\tabularnewline
\hline 
{\footnotesize{}Test B} & {\footnotesize{}82.55} & {\footnotesize{}84.09} & {\footnotesize{}85.91} & {\footnotesize{}85.05} & {\footnotesize{}81.67} & {\footnotesize{}81.53}\tabularnewline
\hline 
{\footnotesize{}Test C} & {\footnotesize{}79.14} & {\footnotesize{}73.01} & {\footnotesize{}82.90} & {\footnotesize{}82.37} & {\footnotesize{}75.79} & {\footnotesize{}75.99}\tabularnewline
\hline 
\emph{\footnotesize{}Overall} & \emph{\footnotesize{}80.57} & \emph{\footnotesize{}81.22} & \emph{\footnotesize{}85.53} & \emph{\footnotesize{}84.85} & \emph{\footnotesize{}79.88} & \emph{\footnotesize{}79.82}\tabularnewline
\hline 
\end{tabular}\protect{\footnotesize \par}}
\end{table}
\begin{table}
\protect\caption{Non-Stereo method on Aurora-4 database\label{tab:Results-on-Aurora4}}

\begin{centering}
{\scriptsize{}}\subfloat[Clean-Training\label{tab:Clean-Training}]{\protect\centering{}{\scriptsize{}}{\footnotesize{}}%
\begin{tabular}{|>{\centering}p{1.8cm}|c|c|c|c|c|c|c|c|c|}
\cline{3-10} 
\multicolumn{1}{>{\centering}p{1.8cm}}{} &  & {\footnotesize{}Clean} & {\footnotesize{}Car} & {\footnotesize{}Babble} & {\footnotesize{}Street} & {\footnotesize{}Restaurant} & {\footnotesize{}Airport} & {\footnotesize{}Station} & {\footnotesize{}Average}\tabularnewline
\hline 
\multirow{2}{1.8cm}{{\footnotesize{}Baseline}} & {\footnotesize{}Mic-1} & {\footnotesize{}87.63} & {\footnotesize{}75.58} & {\footnotesize{}52.77} & {\footnotesize{}52.83} & {\footnotesize{}46.53} & {\footnotesize{}56.38} & {\footnotesize{}45.30} & \multirow{2}{*}{{\footnotesize{}54.73}}\tabularnewline
\cline{2-9} 
 & {\footnotesize{}Mic-2} & {\footnotesize{}77.40} & {\footnotesize{}64.39} & {\footnotesize{}45.15} & {\footnotesize{}42.03} & {\footnotesize{}36.26} & {\footnotesize{}47.69} & {\footnotesize{}36.32} & \tabularnewline
\hline 
\hline 
\multirow{2}{1.8cm}{{\footnotesize{}Non-Stereo Method}} & {\footnotesize{}Mic-1} & {\footnotesize{}87.39} & {\footnotesize{}77.21} & {\footnotesize{}62.79} & {\footnotesize{}59.29} & {\footnotesize{}57.43} & {\footnotesize{}60.9} & {\footnotesize{}57.99} & \multirow{2}{*}{{\footnotesize{}61.78}}\tabularnewline
\cline{2-9} 
 & {\footnotesize{}Mic-2} & {\footnotesize{}78.95} & {\footnotesize{}67.18} & {\footnotesize{}55.41} & {\footnotesize{}50.72} & {\footnotesize{}46.31} & {\footnotesize{}54.55} & {\footnotesize{}48.76} & \tabularnewline
\hline 
\end{tabular}\protect{\footnotesize \par}}
\par\end{centering}{\scriptsize \par}

\centering{}{\scriptsize{}}\subfloat[Multi-Training\label{tab:Multi-Training}]{\protect\centering{}{\scriptsize{}}{\footnotesize{}}%
\begin{tabular}{|>{\centering}p{1.8cm}|c|c|c|c|c|c|c|c|c|}
\cline{3-10} 
\multicolumn{1}{>{\centering}p{1.8cm}}{} &  & {\footnotesize{}Clean} & {\footnotesize{}Car} & {\footnotesize{}Babble} & {\footnotesize{}Street} & {\footnotesize{}Restaurant} & {\footnotesize{}Airport} & {\footnotesize{}Station} & {\footnotesize{}Average}\tabularnewline
\hline 
\multirow{2}{1.8cm}{{\footnotesize{}Baseline}} & {\footnotesize{}Mic-1} & {\footnotesize{}86.89} & {\footnotesize{}82.83} & {\footnotesize{}70.52} & {\footnotesize{}68.52} & {\footnotesize{}65.37} & {\footnotesize{}72.73} & {\footnotesize{}64.28} & \multirow{2}{*}{{\footnotesize{}70.16}}\tabularnewline
\cline{2-9} 
 & {\footnotesize{}Mic-2} & {\footnotesize{}82.29} & {\footnotesize{}77.36} & {\footnotesize{}65.48} & {\footnotesize{}62.23} & {\footnotesize{}57.61} & {\footnotesize{}67.29} & {\footnotesize{}58.77} & \tabularnewline
\hline 
\hline 
\multirow{2}{1.8cm}{{\footnotesize{}Non-Stereo Method}} & {\footnotesize{}Mic-1} & {\footnotesize{}86.21} & {\footnotesize{}82.03} & {\footnotesize{}71.64} & {\footnotesize{}67.59} & {\footnotesize{}66.47} & {\footnotesize{}71.32} & {\footnotesize{}65.89} & \multirow{2}{*}{{\footnotesize{}69.98}}\tabularnewline
\cline{2-9} 
 & {\footnotesize{}Mic-2} & {\footnotesize{}80.59} & {\footnotesize{}75.68} & {\footnotesize{}66.41} & {\footnotesize{}62.11} & {\footnotesize{}58.68} & {\footnotesize{}65.31} & {\footnotesize{}59.85} & \tabularnewline
\hline 
\end{tabular}\protect{\footnotesize \par}}{\scriptsize \par}
\end{table}

Table \ref{tab:Individual-Methods} summarises the results of various
algorithms discussed, on Aurora-2 dataset. All the results are shown
in \% accuracy. All SNRs levels mentioned are in decibels. The first
seven rows report the overall results on all 10 test noise conditions.
The rest of the rows report the average values in the SNR range $20-0$
dB. Table \ref{tab:Run-time-adaptation} shows the results of run-time
adaptation (indicated as RA) using various methods. For reference,
the result of standard MLLR adaptation on HMMs \cite{mllrgales} has
been shown in Table \ref{tab:Run-time-adaptation}, which computes
a global 39 dimensional mean transformation, and uses two-pass Viterbi
decoding. Table \ref{tab:Results-on-Aurora4} shows the experimental
results on Aurora-4 database. Table \ref{tab:Clean-Training} shows
the results of non-stereo method on Aurora-4 database using clean-trained
HMMs. Table \ref{tab:Multi-Training} shows the similar results for
multi-trained HMMs, using the standard multi-training dataset.

It can be seen that M-SPLICE improves over SPLICE at all noise conditions
and SNR levels and gives an \emph{absolute} improvement of $8.6\%$
in test-set C and $2.93\%$ overall. Run-time compensation in SPLICE
framework gives improvements over standard MLLR in test-sets A and
B, whereas M-SPLICE gives improvements in all conditions. Here $9.89\%$
absolute improvement can be observed over SPLICE with run-time noise
adaptation, and $4.96\%$ over standard MLLR. Finally, non-stereo
method, though not using stereo data, shows $10.37\%$ and $7.05\%$
absolute improvements over Aurora-2 and Aurora-4 clean baseline models
respectively, and a slight degradation w.r.t. SPLICE in all test cases.
Run-time noise adaptation results of non-stereo method are comparable
to that of standard MLLR, and are computationally less expensive.
It can be observed that non-stereo method gives performance similar
to that of multi-condition training.

\section{Discussion}

\label{sec:Discussion-1}

In terms of computational cost, the methods M-SPLICE and non-stereo
methods are identical during testing as compared to conventional SPLICE.
Also, there is almost negligible increase in cost during training.
The MLLR mean adaptation in both non-stereo method and run-time adaptation
are computationally very efficient, and do not need Viterbi decoding.
The diagonal versions of the proposed methods give comparable performances.

In terms of performance, M-SPLICE is able to achieve good results
in all cases without any use of adaptation data, especially in unseen
cases. In non-stereo method, one-to-one mixture correspondence is
assumed between noise and clean GMMs. The method gives slight degradation
in performance. This could be attributed to neglecting the outlier
data.

Comparing with other existing feature normalisation techniques, the
techniques in SPLICE framework operate on individual feature vectors,
and no estimation of parameters is required from test data. So these
methods do not suffer from test data insufficiency problems, and are
advantageous for shorter utterances. Also, the testing process is
usually faster, and are easily implementable in real-time applications.
So by extending the methods to non-stereo data, we believe that they
become more useful in many applications.

\chapter{Conclusion and Future Work\label{chap:Conclusion-and-Future}}

In this thesis, feature normalisation methods suitable for noise robust
real-time ASR have been studied. When there is no information about
noise, it has been shown that an additional feature processing block
in the MFCC extraction process can be included for noise-robustness.
The additional block rebuilds all the speech frames from non-negative
linear combinations of the speech subspace basis vectors in the LMFB
domain. The new features are also shown to give improved performance
when used with the existing techniques such as HEQ and HLDA.

In future, the methods may be improved by imposing sparseness constraints
for learning better and more meaningful speech dictionaries. Also,
the current methods involve iterative estimation of weights $\mathbf{H}$
during testing. Though the implementation is simple, owing to its
slow convergence a large number of iterations is required. Addressing
this issue could be another possible future work. The methods may
be implemented in higher dimensional spaces for improvements, possibly
involving DFT of larger number of samples during feature extraction.

In the presence of stereo training data, a modified version of the
SPLICE algorithm has been proposed for noise robust ASR. It is better
compliant with the assumptions of SPLICE, and improves the recognition
in highly mismatched and unseen noise conditions. An extension of
the methods to non-stereo data has been presented. Finally, a convenient
run-time adaptation framework has been explained, which is computationally
much cheaper than the standard MLLR adaptation of HMMs.

In future, the efficiency of the non-stereo extension of SPLICE can
be improved. Better techniques could be proposed to achieve this,
which either give fewer outliers in their mixture distribution plot,
or do not neglect the outlier data. M-SPLICE could also be extended
in uncertainty decoding framework, which has gained popularity \cite{udsplice}
over the conventional SPLICE.

\chapter*{Derivation of SPLICE Parameters}

Let $\mathbf{W}_{m}=\begin{bmatrix}\mathbf{b}_{m} & \mathbf{A}_{m}\end{bmatrix}$
and $\mathbf{y}'=\begin{bmatrix}1\\
\mathbf{y}
\end{bmatrix}$. $\left(\mathbf{x}_{n},\mathbf{y}_{n}\right)$ can be substituted
in 
\[
p(\mathbf{x}\,|\,\mathbf{y},m)\sim\mathcal{N}\left(\mathbf{x}\,;\,\mathbf{A}_{m}\mathbf{y}+\mathbf{b}_{m},\boldsymbol{\Sigma}_{x,m}\right)
\]
to obtain
\[
p\left(\mathbf{x}_{n}\,|\,\mathbf{y}_{n},m\right)=\frac{1}{\left(2\pi\right)^{\frac{D}{2}}\left|\boldsymbol{\Sigma}_{x,m}\right|^{\frac{1}{2}}}e^{-\frac{1}{2}\left(\mathbf{x}-\mathbf{W}_{m}\mathbf{y}_{n}'\right)^{T}\boldsymbol{\Sigma}_{x,m}^{-1}\left(\mathbf{x}-\mathbf{W}_{m}\mathbf{y}_{n}'\right)}
\]

Now, the log-likelihood function can be expanded as 
\[
\mathcal{L}=\log\left[\prod_{n=1}^{N}p\left(\mathbf{x}_{n},\mathbf{y}_{n}\right)\right]=\sum_{n=1}^{N}\log\left[\sum_{m=1}^{M}p(\mathbf{x}_{n},\mathbf{y}_{n},m)\right]
\]
\[
\mathcal{L}=\sum_{n=1}^{N}\log\left[\sum_{m=1}^{M}p\left(\mathbf{x}_{n}\,|\,\mathbf{y}_{n},m\right)p\left(\mathbf{y}_{n}\,|\,m\right)P\left(m\right)\right]
\]

Differentiating w.r.t $\mathbf{W}_{m}$ and equating to zero, and
using the matrix identity 
\[
\frac{\partial}{\partial\mathbf{A}}\left(\mathbf{x}-\mathbf{A}\mathbf{s}\right)\mathbf{B}\left(\mathbf{x}-\mathbf{A}\mathbf{s}\right)=-2\mathbf{B}\left(\mathbf{x}-\mathbf{A}\mathbf{s}\right)\mathbf{s}^{T}
\]
\[
\implies\frac{\partial\mathcal{L}}{\partial\mathbf{W}_{m}}=\sum_{n=1}^{N}\frac{1}{p(\mathbf{x}_{n},\mathbf{y}_{n})}p\left(\mathbf{y}_{n}\,|\,m\right)P\left(m\right)p\left(\mathbf{x}_{n}\,|\,\mathbf{y}_{n},m\right)\boldsymbol{\Sigma}_{x,m}^{-1}\left(\mathbf{x}_{n}-\mathbf{W}_{m}\mathbf{y}_{n}'\right)\mathbf{y}_{n}'^{T}=0
\]
\[
\implies\boldsymbol{\Sigma}_{x,m}^{-1}\left(\sum_{n=1}^{N}p\left(m\,|\,\mathbf{x}_{n},\mathbf{y}_{n}\right)\left(\mathbf{x}_{n}-\mathbf{W}_{m}\mathbf{y}_{n}'\right)\mathbf{y}_{n}'^{T}\right)=0
\]

SPLICE assumes a perfect correlation between $\mathbf{x}$ and $\mathbf{y}$,
so 
\[
p\left(m\,|\,\mathbf{x}_{n},\mathbf{y}_{n}\right)\approx p\left(m\,|\,\mathbf{x}_{n}\right)\approx p\left(m\,|\,\mathbf{y}_{n}\right)
\]

Since $\Gamma_{m}^{-1}$ is non-singular and the above matrix product
is zero, it can be proved that
\[
\sum_{n=1}^{N}p\left(m\,|\,\mathbf{y}_{n}\right)\left(\mathbf{x}_{n}-\mathbf{W}_{m}\mathbf{y}_{n}'\right)\mathbf{y}_{n}'^{T}=0
\]

Solving for $\mathbf{W}_{m}$ yields
\[
\mathbf{W}_{m}=\left[\sum_{n=1}^{N}p\left(m\,|\,\mathbf{y}_{n}\right)\mathbf{x}_{n}\mathbf{y}_{n}'^{T}\right]\left[\sum_{n=1}^{N}p\left(m\,|\,\mathbf{y}_{n}\right)\mathbf{y}_{n}'\mathbf{y}_{n}'^{T}\right]^{-1}
\]

\chapter*{Publications Based on this Thesis}
\begin{enumerate}
\item D. S. Pavan Kumar, N. Vishnu Prasad, Vikas Joshi, S. Umesh, \textquotedblleft Modified
SPLICE and its Extension to Non-Stereo Data for Noise Robust Speech
Recognition,\textquotedblright{} accepted in \emph{IEEE Automatic
Speech Recognition and Understanding Workshop}, Olomouc, Dec. 2013.
(\href{http://dx.doi.org/10.1109/ASRU.2013.6707725}{\textcolor{blue}{View
Online}})\bigskip{}

\item D. S. Pavan Kumar, Raghavendra Bilgi and S. Umesh, \textquotedblleft Non-Negative
Subspace Projection During Conventional MFCC Feature Extraction for
Noise Robust Speech Recognition,\textquotedblright{} \emph{National
Conference on Communications}, New Delhi, Feb. 2013. (\href{http://dx.doi.org/10.1109/NCC.2013.6487993}{\textcolor{blue}{View
Online}})
\end{enumerate}
\bibliographystyle{apalike}
\bibliography{Biblio}

\end{document}